%% file: main.tex
\PassOptionsToPackage{svgnames,table}{xcolor}

\documentclass[11pt,letterpaper]{mystyle}

\usepackage[comma,authoryear,compress]{natbib}
\usepackage{xcolor}
\usepackage{hyperref}
\usepackage[all]{hypcap}
\usepackage{cleveref}

\hypersetup{
    colorlinks=true,
    citecolor=YaleBlue,
}
\usepackage{algorithm}
\usepackage{algpseudocode}

\usepackage{amsmath}
\usepackage{amssymb}
\usepackage{amsfonts}
\usepackage{mathtools}
\usepackage{amsthm}
\usepackage{mathrsfs}
\usepackage{nicefrac}
\usepackage{dsfont}
\usepackage{fdsymbol}

\usepackage{graphicx}
\usepackage{subcaption}
\usepackage{float}
\usepackage{wrapfig}
\usepackage{rotating}
\usepackage{adjustbox}
\usepackage{capt-of}

\usepackage{booktabs}
\usepackage{bigstrut}
\usepackage{threeparttable}
\usepackage{multirow}
\usepackage{tabularx}
\usepackage{array}
\usepackage{makecell}

\usepackage[font=small,labelfont=bf]{caption}

\usepackage[utf8]{inputenc}
\usepackage[T1]{fontenc}
\usepackage{url}
\usepackage{microtype}
\usepackage{enumitem}
\usepackage{lipsum}
\usepackage{stackengine}
\usepackage{bxcoloremoji}

\input{macro}

\newtcolorbox{AIbox}[2][]{aibox,title=#2,#1}

\definecolor{lightblue}{rgb}{0.22,0.45,0.70}
\definecolor{Gray}{gray}{0.95}
\definecolor{Cornsilk}{rgb}{1.0,0.97,0.86}

\definecolor{setablue}{RGB}{220,232,246}
\definecolor{tmaxgreen}{RGB}{224,239,220}
\definecolor{blockgray}{RGB}{245,245,245}
\definecolor{bestbg}{RGB}{198,224,255}
\definecolor{secondbg}{RGB}{255,222,173}

\definecolor{refshade}{RGB}{245,245,245}
\definecolor{usershade}{RGB}{239,246,255}
\definecolor{adaptshade}{RGB}{255,247,237}
\definecolor{selfshade}{RGB}{240,253,244}
\definecolor{bilatshade}{RGB}{248,245,255}
\definecolor{atoashade}{RGB}{240,247,250}

\makeatletter
\renewcommand{\AB@affilsepx}{\protect\\\protect\Affilfont}
\makeatother

\tcbuselibrary{skins,breakable}
\newtcolorbox{cfexample}[1]{
  enhanced,
  breakable,
  colback=white,
  colframe=black,
  colbacktitle=black,
  coltitle=white,
  fonttitle=\bfseries,
  title={#1},
  arc=2mm,
  boxrule=0.8pt,
  left=1.2mm,
  right=1.2mm,
  top=1.0mm,
  bottom=1.0mm
}

\newtcolorbox{promptbox}[1]{
  colback=white,
  colframe=black,
  coltitle=white,
  colbacktitle=black,
  title=\textbf{#1},
  fonttitle=\bfseries,
  boxrule=0.8pt,
  arc=2mm,
  left=2mm,
  right=2mm,
  top=1mm,
  bottom=1mm
}

\title{CIPO: Counterfactual Imagination Policy Optimization for Adaptive Tool Granularity Selection}

\author{
  Yu Li$^1$, Yunlu Wan$^1$, Zijian Zhu$^1$, Han Luo$^1$, Chao Ren$^2$, Long-Fei Li$^3$, and Lei Feng
}

\affil[1]{School of Computer Science and Engineering,
Southeast University, Nanjing, China}

\affil[2]{KTH Royal Institute of Technology 
\quad \textsuperscript{3}Huawei Noah's Ark Lab, China}

\correspondingauthor{Lei Feng, fenglei@seu.edu.cn}

\begin{document}

\begin{abstract}
{\centering\textbf{Abstract}\par}
\vspace{0.5em}
Large language model (LLM) agents solve complex tasks through multi-step interactions with external tools. These interactions often contain recurring local tool sequences. Treating such sequences as composite ``Skills'' can shorten tool-use trajectories and reduce repeated low-level decisions.
However, when atomic tools and composite skills coexist, skill use becomes a policy problem: the agent must decide whether the current state requires atomic fine control or skill-level abstraction. 
In this paper, we argue that effective skill use should be studied as adaptive tool granularity selection. The most direct training signal for this problem is to compare the consequences of atomic and skill choices available from the same state. 
Based on this view, we propose CIPO, a \textbf{C}ounterfactual \textbf{I}magination \textbf{P}olicy \textbf{O}ptimization framework for adaptive tool granularity. CIPO constructs executable skills through budget-constrained mining of successful tool-use trajectories and trains granularity decisions with counterfactual branch rollouts.
For each base rollout, CIPO branches at the first eligible granularity decision and replaces the chosen action with a feasible atomic or skill alternative. The paired outcome difference serves as a supplementary reward for policy optimization. 
Experiments across multiple benchmarks and model backbones show that CIPO improves task success and decision efficiency over baselines. 
Further analyses show that CIPO learns effective skill use by improving the choice between atomic tools and composite skills based on the current state, without simply increasing skill frequency.
\end{abstract}

\maketitle
\vspace{3mm}

\section{Introduction}

Large language model (LLM) agents use external tools to solve complex tasks involving information retrieval \citep{nakano2021webgpt, min, min2}, code execution \citep{gao2023pal}, file manipulation \citep{chen2023fireact}, and API interaction~\citep{patil2024gorilla, react, schick2023toolformer, API-Bank, ToolLLM, zeng2024agenttuning, li2026phgpo, mateval}. As these tasks become more complex, tool use often unfolds as long trajectories in which the agent repeatedly selects tools under changing intermediate states \citep{liu2023agentbench, zhou2024webarena, tau, lu2025toolsandbox, toolathlon}. Such trajectories commonly contain reusable local routines, including metadata retrieval followed by content reading, question parsing followed by answer extraction, and directory listing followed by file reading. These recurring sequences indicate that long-horizon tool use is not merely a chain of isolated tool invocations. It also contains reusable procedures that can be packaged as composite ``Skills'' \citep{sutton1999between, liang2023code, wang2023voyager, jiang2026sok}, allowing several low-level tool invocations to be compressed into a single higher-level action.

Although recent work has explored different forms of reuse processes in LLM agents, existing approaches do not fully address how these reusable procedures should be constructed, executed, and selected during policy learning. Memory-based methods retrieve past trajectories, reflections, or task records to guide the current agent~\citep{park2023generative, shinn2023reflexion, zhao2024expel, packer2023memgpt}, but the retrieved experience usually remains in the context and does not change the action space \citep{xu2025mem, chhikara2025mem0, fang2025memp}. 
Workflow reuse methods store or synthesize reusable procedures from previous tasks~\citep{wang2025agent, chen2024automanual, li2024autoflow, alfow}, but these procedures are often treated as fixed guidance rather than actions that compete with atomic tools at each decision state. Tool merging and compilation methods combine tools or tool invocations into higher-level interfaces~\citep{awo, llmcompiler, llmtoolcompiler}, but direct merging can create a large number of overlapping candidates. 
Recent skill-learning methods further suggest that tool-use procedures can be reused as skills~\citep{xia2026skillrl, lu2026skill0, ni2026trace2skill, jiang2026sok}, yet they often leave open the question of how an agent should decide whether the current state calls for an atomic tool or a composite skill. These limitations suggest that reusable tool procedures introduce not only a construction problem, but also an execution problem and a policy problem.

\begin{figure}
  \centering
  \includegraphics[width=\linewidth]{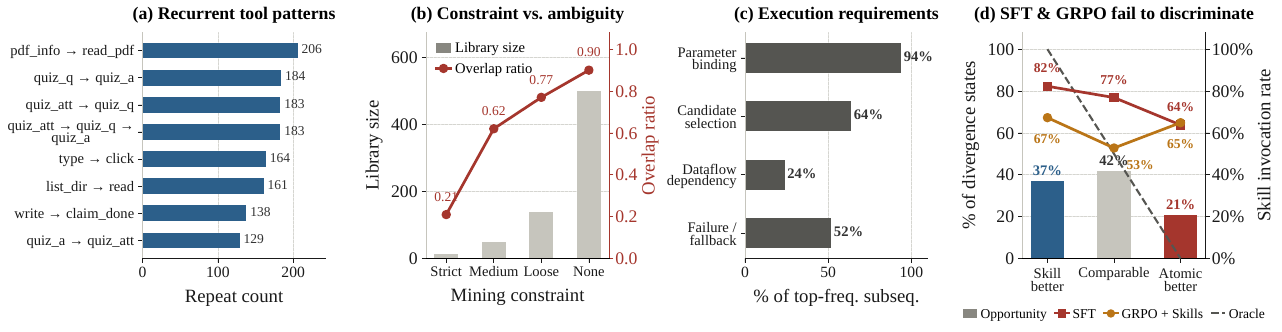}
    \caption{
    Pilot analyses on tool-use granularity.
    (a) Successful TOOLATHLON trajectories contain recurrent tool patterns.
    (b) Looser mining constraints increase both the skill library size and candidate overlap, motivating budget-constrained skill construction.
    (c) Frequent subsequences often require runtime execution support, including parameter binding, candidate selection, dataflow handling, and failure recovery.
    (d) Granularity preference varies across states, while SFT and ``GRPO + Skills" fail to follow the oracle trend, suggesting the need for explicit granularity training.
    }
    \label{fig:motivation}
  \vskip -0.6cm
\end{figure}

To examine these challenges, we conduct a series of pilot analyses. As shown in Figure~\ref{fig:motivation}(a), recurrent tool patterns are common in successful TOOLATHLON~\citep{toolathlon} trajectories, with several subsequences appearing over 100 times. This result supports the basic premise that skill construction can exploit repeated local structure. However, Figure~\ref{fig:motivation}(b) shows that relaxing mining constraints rapidly increases both the size of the skill library and the overlap among mined candidates. This pattern indicates that unconstrained tool merging can produce an ambiguous action space, where many candidates express similar procedures. Figure~\ref{fig:motivation}(c) further shows that frequent subsequences are rarely executable as fixed pipelines. Most of them require runtime decisions, including parameter binding, candidate selection, dataflow handling, or failure recovery. Finally, Figure~\ref{fig:motivation}(d) shows that granularity preference is state-dependent. Skills are better in 37\% of the examined states, atomic tools are better in 21\%, and the remaining states are comparable. However, SFT and ``GRPO + Skills" do not clearly separate these categories. Their skill invocation rates remain far from the oracle trend, which suggests that standard training can expose the model to skills but does not provide sufficient feedback for state-dependent granularity selection.

These findings imply three requirements for skill-augmented tool use. First, skill construction should be constrained, because frequent subsequences must be selected without causing uncontrolled library growth or severe overlap. Second, mined subsequences must be instantiated as executable skills, because useful skills need parameter interfaces, selection control, execution traces, and interruption handling rather than fixed symbolic pipelines. Third, policy training must teach the agent when to use each granularity. A composite skill can improve efficiency when its internal procedure matches the current subtask, but it can also hide useful intermediate observations or execute unnecessary steps when finer control is needed. We formulate this third requirement as adaptive tool granularity selection \citep{ning2024wtu}. Given a decision state where an atomic action and a composite skill are feasible, the policy should choose the granularity that better supports the remaining trajectory. The most direct training signal for this choice is a comparison between the consequences of atomic and skill actions available from the same state \citep{prm, liu2025cart}.

Based on these requirements, we propose CIPO, a \textbf{C}ounterfactual \textbf{I}magination \textbf{P}olicy \textbf{O}ptimization framework for adaptive tool granularity. CIPO first constructs a compact set of executable skills from successful tool-use trajectories through budget-constrained BPE-style \citep{bep} mining and trace-aware instantiation. The mining stage selects recurring tool subsequences under explicit constraints, while the instantiation stage converts each selected subsequence into a callable skill with parameter binding, selection control, execution tracing, and soft interruption. This design preserves atomic tools for fine control while adding composite skills for reusable procedures.

CIPO then extends GRPO \citep{shao2024deepseekmath} to train the policy to choose between atomic tools and composite skills. For each base rollout, CIPO creates a counterfactual branch at the first eligible granularity decision in the base rollout and replaces the chosen action with a feasible atomic or skill alternative. The branch continues under the same policy, which keeps the comparison tied to the same decision context. The paired outcome difference is used as a supplementary reward in GRPO, while the original task reward remains the main training signal. 

We evaluate CIPO across multiple tool-use benchmarks with Qwen2.5-7B and Llama3.1-8B backbones. The results show that CIPO improves task success and decision efficiency over baselines. Further analyses show that the improvement does not come from simply increasing skill frequency. Instead, CIPO learns more effective skill use by improving the policy's ability to select the appropriate tool granularity under the current state.

Our contributions are summarized as follows:

\begin{itemize}[leftmargin=0.5cm,topsep=0pt]
    \item We formulate adaptive tool granularity selection as a key policy problem for skill-augmented LLM agents, showing through pilot analyses that SFT and GRPO with access to skills do not reliably decide when to use atomic tools or composite skills based on their relative benefits in the current state.
    \item We introduce a budget-constrained BPE-style skill construction pipeline that turns recurring tool-use patterns into executable skills with parameter binding, execution traces, and soft interruption.
    \item We propose CIPO, a counterfactual policy optimization framework that improves task success, decision efficiency, and effective skill use across multiple benchmarks and backbones.
\end{itemize}

\section{Related Work}

\noindent\textbf{Tool-Augmented LLM Agents.}
Recent LLM agents use external tools to access information, execute code, and call real-world APIs \citep{nakano2021webgpt, API-Bank, zeng2024agenttuning, yang, yang2026srpo, c3tg, yang2026sca}. ReAct interleaves reasoning with environment actions~\citep{react}, while Toolformer trains language models to decide when and how to call APIs through self-supervised data construction~\citep{schick2023toolformer}. ToolLLM and Gorilla further scale tool learning to large API collections and evaluate models on tool selection and invocation~\citep{ToolLLM,patil2024gorilla}. These works establish tool use as a core interface for LLM agents, but they usually keep individual tool invocations as the basic action unit. We instead consider composing recurring tool sequences into skills and study when a policy should use atomic tools or composite skills.

\noindent\textbf{Experience and Workflow Reuse.}
Another line of work improves agents by reusing past experience as memory, reflection, or procedural guidance. ExpeL extracts lessons from previous tasks and recalls them during later decision making~\citep{zhao2024expel}. MemP and Mem0 study procedural or long-term memory for agent experience reuse~\citep{fang2025memp,chhikara2025mem0}. AWM induces workflows from previous web tasks, and AutoManual builds instruction manuals through interaction with environments~\citep{wang2025agent,chen2024automanual}. These methods show that prior trajectories contain reusable knowledge, but context-level reuse does not reduce the number of low-level tool decisions. It also leaves open when a reusable procedure should be executed as an action rather than only used as guidance.

\noindent\textbf{Tool Composition and Skill Construction.}
Tool composition methods aim to organize multiple tool invocations into higher-level execution units. LLMCompiler plans and executes multiple function calls to improve orchestration efficiency~\citep{llmcompiler}, while LLM-Tool Compiler and AWO merge or compile recurring tool operations into more efficient interfaces~\citep{llmtoolcompiler,awo}. Recent skill-oriented methods, including SkillRL, SKILL0, and Trace2Skill, further explore how reusable skills can be built or distilled from agent experience~\citep{xia2026skillrl,lu2026skill0,ni2026trace2skill}. These methods move beyond context-level reuse, but they often focus on constructing or compiling skills, leaving skill-library ambiguity and atomic-versus-skill selection underexplored.

\noindent\textbf{Policy Optimization for Tool-Use Agents.}
Reinforcement learning has been used to improve multi-turn tool-use agents, where training signals are often defined from final task outcomes. GRPO optimizes policies with group-relative advantages~\citep{shao2024deepseekmath}, and recent work extends outcome-based policy optimization to multi-turn tool-calling agents~\citep{li2026mtdrive, cai2026beyond}. Other methods introduce step-level credit assignment for multi-step agent rollouts~\citep{wang2026rtmc, gtool, toolstar, steptool, toolrl}. These methods improve learning from long interactions, but they usually assume a fixed action granularity. CIPO instead targets the granularity decision between atomic tools and composite skills.


\section{Method}

\subsection{Problem Formulation}

We consider a tool-use task with an input instruction \(x\). At step \(t\), the agent observes the interaction history \(h_t=(x,a_1,o_1,\ldots,a_{t-1},o_{t-1})\), where \(a_i\) is a selected action and \(o_i\) is the observation returned by the environment. The policy \(\pi_\theta\) samples the next action as \(a_t\sim\pi_\theta(\cdot\mid h_t)\). A rollout \(\tau=(x,a_1,o_1,\ldots,a_T,o_T)\) ends when the agent produces a final answer or reaches a maximum interaction length, and receives a task reward \(R_{\mathrm{task}}(\tau)\).

Let \(\mathcal{A}_{\mathrm{atom}}\) denote the set of atomic tools. Each atomic tool corresponds to a single executable operation, such as retrieving a file, reading a document, running a command, or querying an API. We also define a set of composite skills \(\mathcal{A}_{\mathrm{skill}}\). Each skill \(s\in\mathcal{A}_{\mathrm{skill}}\) is exposed to the policy as one callable action, while its internal execution follows a sequence of atomic tools \(s=(u_1,\ldots,u_m)\), where \(u_i\in\mathcal{A}_{\mathrm{atom}}\). The augmented action space is
\begin{equation}
\mathcal{A}=\mathcal{A}_{\mathrm{atom}}\cup\mathcal{A}_{\mathrm{skill}}.
\label{eq:augmented_action_space}
\end{equation}

This augmented action space creates a granularity choice. An atomic tool provides fine control through a single operation, whereas a composite skill provides a higher level action that executes a reusable procedure. We denote the granularity of an action by \(g(a)\in\{\mathrm{atom},\mathrm{skill}\}\). A granularity comparison is available at a state \(h_t\) when the same history supports both a feasible atomic action and a related composite skill. In this case, the two actions provide alternative ways to advance the same local subtask at different granularities.

Adaptive tool granularity selection asks the policy to decide, under the current history, whether to take a single atomic step or invoke a composite skill that executes several atomic steps. This is not a global preference for one type of action. An atomic tool is useful when the agent needs intermediate observations or fine control, whereas a skill is useful when its internal procedure matches the current subtask. The policy therefore needs evidence about which granularity works better from the same decision context. CIPO provides this evidence by constructing executable skills and training the policy with counterfactual comparisons between feasible atomic and skill choices.

\subsection{Budget-constrained Skill Mining}
\label{sec:method3.2}

Mining reusable procedures from tool trajectories requires explicit constraints, because frequent-subsequence mining can easily produce redundant candidates. For example, a repeated routine such as \texttt{search\_file} \(\rightarrow\) \texttt{read\_file} \(\rightarrow\) \texttt{extract\_answer} may yield both its full sequence and shorter overlapping subsequences as candidate skills. These variants describe related routines but compete as separate actions, which increases the ambiguity of skill selection. We therefore mine a compact set of recurring patterns under explicit budget constraints.

Given a successful rollout \(\tau\), we convert its atomic tool invocations into an abstract token sequence. Each tool invocation is represented by its tool name and parameter-key signature, written as \(z_t=\mathrm{name}(a_t)[\mathrm{keys}(a_t)]\). This abstraction preserves the structure of a tool invocation while removing task-specific parameter values. For instance, two calls to the same file-reading tool with different file paths share the same token if they require the same parameter keys. We define \(\mathcal{D}_{\mathrm{succ}}\) denote the set of token sequences extracted from successful rollouts.

Inspired by byte-pair encoding in tokenization, which repeatedly merges frequent adjacent symbols, we apply a BPE-style merging procedure to \(\mathcal{D}_{\mathrm{succ}}\). At each round, the algorithm counts adjacent token pairs and selects the most frequent valid pair:
\begin{equation}
(a^\star,b^\star)=\arg\max_{(a,b)}
\sum_{\mathbf{z}\in\mathcal{D}_{\mathrm{succ}}}
\sum_{t=1}^{|\mathbf{z}|-1}
\mathbb{I}[z_t=a, z_{t+1}=b].
\label{eq:bpe_pair_selection}
\end{equation}
All occurrences of \((a^\star,b^\star)\) are merged into a new composite token. The merge process is constrained by the maximum number of merge rounds \(K\), the minimum pair frequency \(f_{\min}\), and the maximum pattern length \(L_{\max}\). These constraints bound the number of generated candidates by \(K\) and the length of each retained candidate by \(L_{\max}\), preventing the skill library from growing with all frequent subsequences in the corpus. We provide implementation details and the values used in our experiments in Appendix~\ref{app:mining_details}.

The output of this stage is a compact set of symbolic skill candidates. Each candidate stores its merged token, its atomic tool sequence, its frequency statistics, and its merge history. These candidates identify recurring tool-use structures, but they are not yet executable skills. The next stage converts them into callable actions with explicit parameter interfaces and runtime execution support.

\begin{figure}[t]
  \centering
  \includegraphics[width=\linewidth]{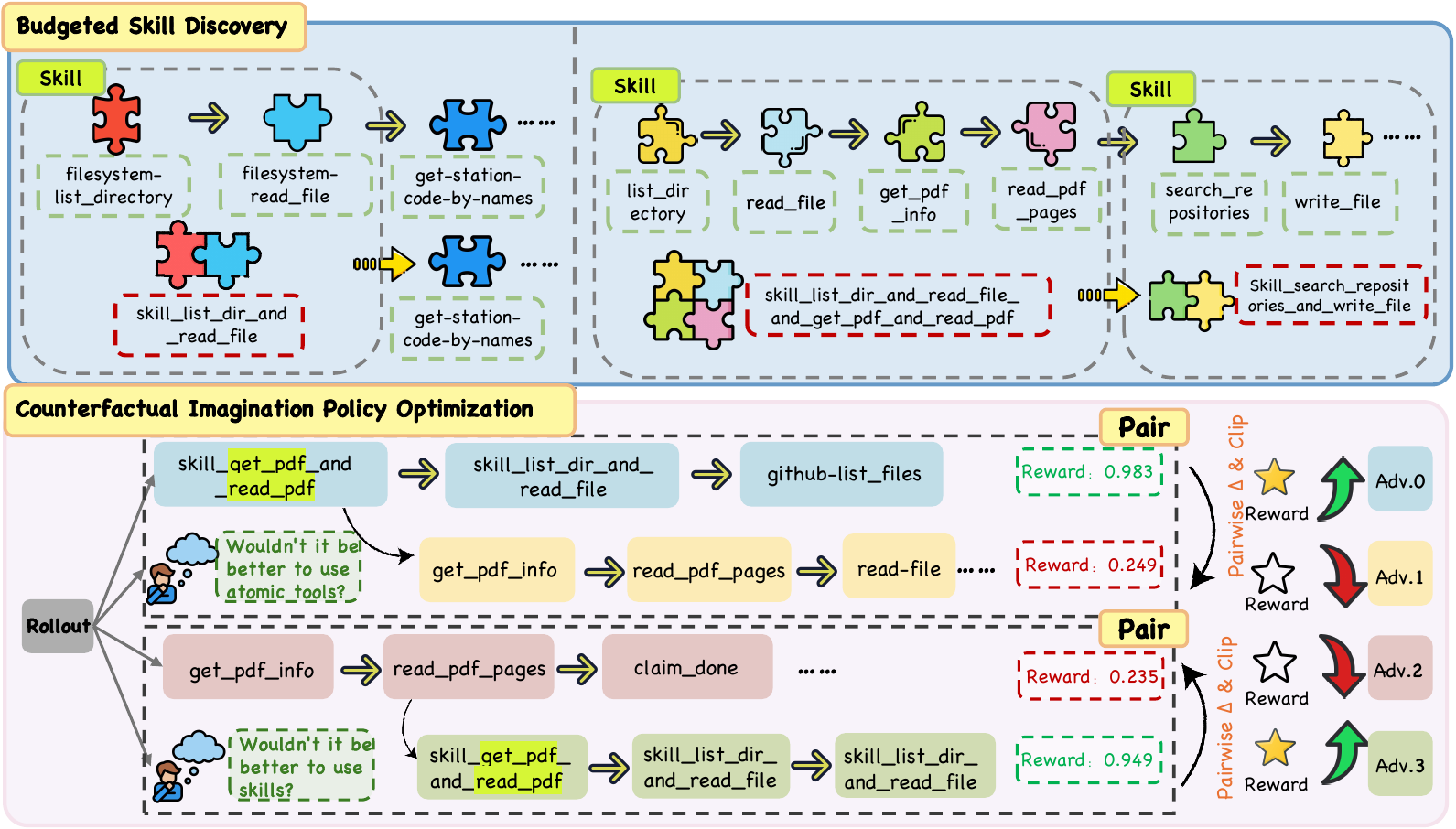}
\caption{
Overview of CIPO. Budgeted skill discovery mines recurring tool-use patterns and instantiates executable skills. Counterfactual imagination policy optimization creates paired branch rollouts by replacing the first eligible granularity choice with an atomic or skill alternative. The clipped reward difference provides a supplementary signal for adaptive tool granularity.
}
    \label{fig:method}
\end{figure}

\subsection{Executable Skill Instantiation}

The mining stage returns symbolic skill candidates, each represented as an ordered sequence of atomic tools. However, a symbolic sequence cannot be directly used by the policy. To make a candidate executable, CIPO registers it as a callable skill with a tool schema, an execution plan, and runtime control fields. For a candidate \(s=(u_1,\ldots,u_m)\), the registered skill exposes a single action to the policy, while internally preserving the ordered execution of its atomic tools.

A central issue is parameter assignment. Some parameters must be provided by the policy because they depend on the current task state, while others can be passed internally from earlier tool outputs. CIPO separates these two cases through parameter analysis. Parameters required by the first atomic tool are exposed by default. For later tools, a parameter is treated as internal if it can be matched to an output field from a previous step; otherwise, it remains exposed to the policy. This design keeps the external skill interface compact while preserving the dataflow needed for execution.

CIPO also assigns each skill an execution template. Most candidates use a sequential template, where atomic tools are executed in order with internal parameter passing. If an intermediate tool returns multiple candidates and a later tool requires a target item, the skill uses a selection template. In this case, the skill exposes a selection control parameter, allowing the policy to decide how to choose from returned candidates. This prevents the skill from becoming a fixed pipeline and keeps the policy involved in runtime decisions.

During execution, CIPO records a trace for every internal step, including the atomic tool name, arguments, returned observation, and execution status. If an internal step fails or cannot be completed, the skill performs a soft interruption. It returns the partial trace and the available intermediate result, then hands control back to the policy. This trace-aware design makes composite skills inspectable and recoverable, rather than opaque actions that either fully succeed or fail without feedback.

\subsection{CIPO: Counterfactual Imagination Policy Optimization}

After the augmented action space is built, the remaining challenge is to train the policy to choose the appropriate granularity under the current context. Applying GRPO in the augmented action space provides rollout-level rewards but does not directly compare atomic and skill actions from the same decision state. CIPO adds such a comparison by augmenting GRPO with counterfactual branch rollouts. The original task reward remains the main optimization signal, while the counterfactual branch provides a supplementary signal for granularity selection.

\paragraph{Base rollout generation.}
For each training instruction \(x\), CIPO samples base rollouts from the current policy over the augmented action space \(\mathcal{A}\). Each base rollout is a complete interaction that may contain both atomic tool invocations and composite skill calls. To match the rollout budget of GRPO + Skills, CIPO fixes the total number of rollouts per instruction to $G$. Let \(G_{\mathrm{base}}\) denote the number of sampled base rollouts and \(G_{\mathrm{cf}}\) denote the number of counterfactual branches. The rollout budget satisfies
\begin{equation}
G_{\mathrm{base}} + G_{\mathrm{cf}} = G .
\label{eq:rollout_budget}
\end{equation}
In our implementation, we use \(G=4\): CIPO first samples two base rollouts and then constructs at most one counterfactual branch for each base rollout. If a base rollout does not contain a decision state where an alternative granularity is available, its reserved branch budget is filled with an additional base rollout. These additional base rollouts participate in the standard GRPO group normalization, but only valid base-branch pairs receive the granularity-specific supplementary reward.

For each base rollout, CIPO stores the interaction history together with the selected actions, action positions, and skill execution traces. This information allows a counterfactual branch to restart from the exact prefix before a granularity decision. Given a base rollout \(\tau=(x,a_1,o_1,\ldots,a_T,o_T)\), CIPO scans its action sequence and identifies the first decision state where an alternative action at the other granularity is available.

\paragraph{Counterfactual branching.}
After a base rollout is generated, CIPO constructs a paired counterfactual rollout when an alternative granularity is available. The algorithm scans the base rollout from the beginning and branches at the first decision state where the chosen action can be replaced by a feasible action at the other granularity. If the base action is an atomic tool \(a_t\), CIPO selects a composite skill \(s^\star\) whose internal tool chain contains \(a_t\). If the base action is a skill \(s=(u_1,\ldots,u_m)\), CIPO uses its first internal atomic tool \(u_1\) as the alternative action. We denote the alternative action by \(\tilde{a}_t\). Detailed replacement rules and examples are provided in Appendix~\ref{app:cf_replacement}.

The counterfactual branch starts from the same prefix \(h_t\) as the base rollout. CIPO executes \(\tilde{a}_t\), appends the returned observation to the history, and lets the same policy complete the remaining interaction. The resulting branch is denoted as \(\tau_{\mathrm{cf}}\), while the original rollout is denoted as \(\tau_{\mathrm{base}}\). This construction compares two granularity choices from the same decision context and evaluates their effect through the subsequent trajectory. CIPO branches at most once for each base rollout, which keeps the paired comparison clean and avoids an exponential growth of counterfactual trajectories.

We branch only at the first such decision in each base rollout. This design keeps the base rollout and the branch identical before the replacement and changes only one granularity choice, making the paired reward difference easier to interpret. It also avoids the additional rollout cost and mixed interventions that would arise from branching at every eligible step. Across training, different prompts and sampled rollouts expose different branch points.

\paragraph{Counterfactual supplementary reward.}
The branch comparison provides a direct signal for the granularity decision. Let \(R_0(\tau)\) denote the base rollout reward computed from task progress, skill behavior, and execution efficiency. The detailed reward composition is described in Appendix~\ref{app:reward_composition}. For a valid base-branch pair, CIPO computes the outcome difference
\begin{equation}
\Delta = R_0(\tau_{\mathrm{cf}}) - R_0(\tau_{\mathrm{base}}).
\label{eq:cf_delta}
\end{equation}
A positive \(\Delta\) means that the alternative granularity leads to a better rollout, while a negative \(\Delta\) means that the original choice is better. To prevent this signal from dominating the task reward, CIPO scales and clips the difference:
\begin{equation}
r_{\mathrm{cf}} =
\mathrm{clip}\left(\alpha \Delta,\,-\epsilon_{\mathrm{cf}},\,\epsilon_{\mathrm{cf}}\right),
\label{eq:cf_reward}
\end{equation}
where \(\alpha\) controls the strength of the counterfactual signal and \(\epsilon_{\mathrm{cf}}\) sets its maximum magnitude.

CIPO first applies this signal symmetrically to the paired rollouts:
\begin{equation}
\hat{R}(\tau_{\mathrm{cf}})=R_0(\tau_{\mathrm{cf}})+r_{\mathrm{cf}},
\qquad
\hat{R}(\tau_{\mathrm{base}})=R_0(\tau_{\mathrm{base}})-r_{\mathrm{cf}}.
\end{equation}
This reward adjustment encourages the policy to prefer the granularity that performs better under the matched decision context. The counterfactual term is used only as a bounded supplementary signal. The original task reward remains the main signal for policy optimization.

\paragraph{Group-relative policy update.}
After reward adjustment, CIPO groups all rollouts for the same instruction, including base rollouts, counterfactual branches, and additional base rollouts used to fill the fixed budget. All adjusted rewards are clipped to be non-negative, \(\hat{R}_i \leftarrow \max\{0,\hat{R}_i\}\), then normalized within the group:
\begin{equation}
\hat{A}_i =
\frac{\hat{R}_i-\mathrm{mean}(\{\hat{R}_j\}_{j=1}^{G})}
{\mathrm{std}(\{\hat{R}_j\}_{j=1}^{G})+\eta}.
\label{eq:group_advantage}
\end{equation}
CIPO then optimizes the policy with the standard GRPO objective using $\hat{A}_i$ as the rollout-level advantage. The core GRPO update remains unchanged; CIPO adds matched counterfactual branches to the rollout group and a bounded granularity-specific signal to the reward.

At inference time, the model observes both atomic tools and composite skills and directly selects actions from $\mathcal{A}$ without counterfactual branching. These branches are used only during training to improve granularity selection. Algorithm~\ref{alg:cipo} summarizes the training procedure.

Overall, CIPO mines recurring patterns from successful trajectories, instantiates them as executable skills, and combines these skills with atomic tools to form the augmented action space. The model is first supervised fine-tuned on successful trajectories in this space to learn the tool-call format and both action granularities. It is then optimized with counterfactual branch rollouts, where the task reward remains the main signal and paired outcome differences provide bounded supplementary signals for granularity selection. At inference time, the policy directly selects actions from the augmented space without counterfactual branching.

\section{Experiments}

\subsection{Experimental Setup}

\paragraph{Benchmarks and models.}
We evaluate CIPO on TOOLATHLON \citep{toolathlon}, TOUCAN \citep{toucan}, and TRAJECT-Bench \citep{traject}, which cover long-horizon tool-use tasks involving retrieval, file access, structured processing, and API-style operations. We use two open-source backbones, Qwen2.5-7B \citep{qwen2} and Llama3.1-8B \citep{llama}, to test whether the method generalizes across model families.

\paragraph{Baselines.}
We compare CIPO with four groups of baselines: general tool-use methods, including ReAct \citep{react} and ToolLLM \citep{ToolLLM}; memory-based methods, including ExpeL \citep{zhao2024expel}, MemP \citep{fang2025memp}, and Mem0 \citep{chhikara2025mem0}; workflow reuse methods, including AWM \citep{wang2025agent}; and tool composition or skill-based methods, including AWO \citep{awo}, LLMCompiler \citep{llmcompiler}, LLM-Tool Compiler \citep{llmtoolcompiler}, SkillRL \citep{xia2026skillrl}, SKILL0 \citep{lu2026skill0}, and Trace2Skill \citep{ni2026trace2skill}. We also compare with SFT \citep{sft}, standard GRPO \citep{shao2024deepseekmath}, and ``GRPO + Skills" in the ablation study.

\paragraph{Metrics.}

Tool F1 measures the order-aware match between predicted and reference tool trajectories after tool-name normalization. TSR is the mean over five independent LLM-judge evaluations of 100 episodes per setting, counting both \texttt{yes} and \texttt{partial} as success. Decision efficiency is measured by \(\mathrm{Comp.}/\mathrm{Raw}\), the ratio of compressed policy decisions to expanded atomic calls. A smaller ratio indicates stronger trajectory compression.



\subsection{Main Results}

\begin{table*}[htbp]
\centering
\caption{
Main results across three benchmarks with Qwen2.5-7B and Llama3.1-8B.
We report Tool F1 (\%) and Task Success Rate (TSR, \%).
TSR is averaged over five independent judge evaluations of the same 100 evaluation episodes in each setting. 
Avg.Comp./Raw is the average compressed decision ratio for composite-action methods; lower values indicate stronger compression.
}
\label{tab:main_results}

\resizebox{\textwidth}{!}{
\begin{tabular}{lccccccccccccc}
\toprule
\multirow{3}{*}{Method}
& \multicolumn{4}{c}{TOOLATHLON}
& \multicolumn{4}{c}{TOUCAN}
& \multicolumn{4}{c}{TRAJECT-Bench}
& \multirow{3}{*}{\begin{tabular}{@{}c@{}}Avg.\\Comp./Raw\(\downarrow\)\end{tabular}} \\
\cmidrule(lr){2-5}
\cmidrule(lr){6-9}
\cmidrule(lr){10-13}

& \multicolumn{2}{c}{Qwen-7B}
& \multicolumn{2}{c}{Llama-8B}
& \multicolumn{2}{c}{Qwen-7B}
& \multicolumn{2}{c}{Llama-8B}
& \multicolumn{2}{c}{Qwen-7B}
& \multicolumn{2}{c}{Llama-8B}
& \\
\cmidrule(lr){2-3}
\cmidrule(lr){4-5}
\cmidrule(lr){6-7}
\cmidrule(lr){8-9}
\cmidrule(lr){10-11}
\cmidrule(lr){12-13}

& F1\(\uparrow\) & TSR\(\uparrow\)
& F1\(\uparrow\) & TSR\(\uparrow\)
& F1\(\uparrow\) & TSR\(\uparrow\)
& F1\(\uparrow\) & TSR\(\uparrow\)
& F1\(\uparrow\) & TSR\(\uparrow\)
& F1\(\uparrow\) & TSR\(\uparrow\)
& \\
\midrule

\rowcolor{gray!15}
\multicolumn{14}{l}{\textit{General tool-use agents}} \\
ReAct 
& 33.00 & 22.73 & 43.79 & 25.41
& 42.74 & 32.68 & 44.00 & 34.22
& 43.64 & 30.96 & 33.71 & 23.58 & -- \\

ToolLLM
& 21.10 & 11.62 & 35.43 & 20.43
& 26.28 & 17.59 & 33.34 & 27.11
& 50.10 & 31.50 & 54.68 & 32.00
& -- \\

\midrule
\rowcolor{gray!15}
\multicolumn{14}{l}{\textit{Memory and workflow reuse}} \\

ExpeL & 18.87 & 3.50 & 30.25 & 6.20 & 4.28 & 0.97 & 2.88 & 0.65 & 6.85 & 0.50 & 7.02 & 1.75 & -- \\
MemP & 19.96 & 2.45 & 21.87 & 3.78 & 26.29 & 2.20 & 25.20 & 3.05 & 36.96 & 11.05 & 46.37 & 15.65 & -- \\
Mem0 & 20.28 & 12.75 & 36.73 & 18.86 & 31.11 & 20.35 & 44.49 & 28.76 & 40.50 & 29.17 & 49.53 & 30.32 & -- \\
AWM & 15.14 & 6.15 & 22.93 & 8.66 & 4.11 & 1.05 & 7.14 & 0.44 & 6.54 & 1.85 & 12.55 & 3.06 & -- \\

\midrule
\rowcolor{gray!15}
\multicolumn{14}{l}{\textit{Tool composition and skill construction}} \\

AWO
& 24.87 & 15.96
& 31.45 & 19.72
& 21.62 & 20.34
& 24.18 & 18.05
& 27.86 & 17.41
& 30.23 & 15.68
& 0.86 \\

LLMComp.
& 26.92 & 10.64
& 21.37 & 11.28
& 20.84 & 12.53
& 23.16 & 11.76
& 18.45 & 9.12
& 27.38 & 15.04
& 0.93 \\

LLM-ToolComp.
& 18.38 & 11.74
& 20.69 & 14.83
& 25.47 & 16.62
& 28.11 & 15.38
& 23.92 & 11.67
& 32.06 & 18.42
& 0.89 \\

SkillRL & 19.08 & 10.50 & 22.60 & 14.17 & 27.75 & 20.15 & 29.15 & 17.41 & 0.82 & 0.34 & 37.27 & 21.36 & 0.88 \\
SKILL0 & 18.23 & 10.74 & 17.65 & 9.17 & 23.77 & 16.40 & 24.44 & 13.50 & 31.32 & 18.78 & 31.43 & 21.33 & 1.00 \\
Trace2Skill & 12.49 & 7.37 & 13.09 & 6.71 & 23.83 & 15.29 & 24.64 & 13.89 & 31.15 & 20.05 & 34.45 & 18.77 & 0.96 \\

\midrule
\rowcolor{green!8}
\textbf{CIPO (Ours)}
& \textbf{35.17} & \textbf{30.78}
& \textbf{48.32} & \textbf{40.55}
& \textbf{48.55} & \textbf{35.50}
& \textbf{50.51} & \textbf{41.32}
& \textbf{50.99} & \textbf{32.00}
& \textbf{55.86} & \textbf{38.75}
& \textbf{0.845 \(\pm\) 0.03} \\

\bottomrule
\end{tabular}
}
\end{table*}

\paragraph{Comparison across method families.}
Table~\ref{tab:main_results} shows that CIPO achieves the strongest overall performance across all three benchmarks and two backbones. General tool-use agents can invoke external tools, but they still rely on long atomic trajectories. Memory and workflow reuse methods bring useful prior experience, yet their gains remain unstable because reuse is mainly provided as context or guidance. Tool composition and skill construction methods reduce repeated operations, but they do not explicitly train the policy to decide when a skill should replace an atomic tool. CIPO directly targets this granularity choice through counterfactual branch rollouts, achieving the best Tool F1 and TSR in all six benchmark-backbone settings. This consistent improvement suggests that adaptive granularity training helps the policy match useful tool trajectories, complete tasks more reliably, and transfer across different tool-use environments.

\paragraph{Decision compression.}
CIPO also gives the best decision compression among methods with composite actions. Its Avg.Comp./Raw is \(0.845\pm0.03\), corresponding to about a \(15.5\%\) reduction in policy-level decisions after skill expansion. This ratio is lower than existing composition and skill-based baselines, indicating stronger trajectory compression. More importantly, CIPO improves compression while maintaining the strongest overall TSR and competitive Tool F1. 
This indicates that CIPO does not simply invoke more skills, but learns when composite skills can reduce repeated decisions without hurting task completion.
Full metrics and complete results are provided in Appendix~\ref{app:additional_results}, with TSR judge details in Appendix~\ref{app:llm_judge}.

\subsection{Skill Usage and Granularity Selection}

\begin{wrapfigure}{r}{0.49\linewidth}
    \centering
    \includegraphics[width=0.84\linewidth]{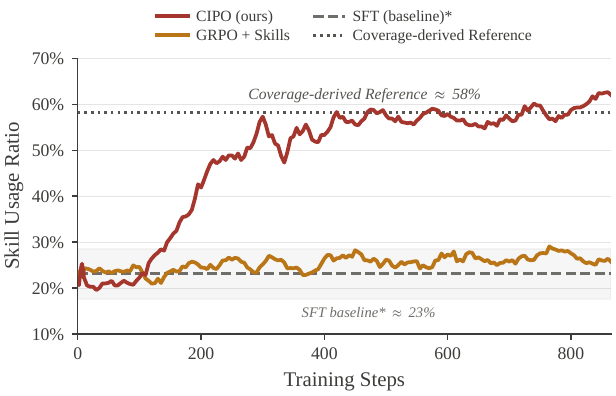}
    \caption{
Skill usage during training. CIPO approaches the coverage-derived reference, while GRPO + Skills remains near its initial level.
    }
    \label{fig:skill_usage}
\end{wrapfigure}

We further examine how skill usage changes during training on TOOLATHLON with Qwen2.5-7B. Figure~\ref{fig:skill_usage} reports the skill usage ratio in online training rollouts, defined as the fraction of policy-level decisions that invoke composite skills. We include a coverage-derived reference level. Given skill-library coverage \(c\) and frequency-weighted average skill length \(m\), the reference ratio is
\begin{equation}
\mathrm{SkillRatio}_{\mathrm{ref}}
=
\frac{c/m}{c/m+(1-c)}.
\label{eq:coverage_reference_main}
\end{equation}

This reference estimates the expected skill usage ratio when replaceable atomic steps are compressed by available skills. It provides a library-level reference and should not be interpreted as a state-level optimal policy. We provide the derivation in Appendix~\ref{app:coverage_reference}.

GRPO + Skills remains close to the SFT baseline, while CIPO increases skill usage toward the coverage-derived reference. This trend reflects training-time skill frequency rather than state-level granularity correctness. Skill-use statistics on final evaluation trajectories come from a different rollout distribution and therefore need not match Figure~\ref{fig:skill_usage}. Table~\ref{tab:ablation} provides complementary evidence: CIPO improves Tool F1, TSR, and decision compression over GRPO + Skills. Together, these results associate altered skill-use behavior with better task performance and efficiency.

\subsection{Ablation Studies}

Table~\ref{tab:ablation} reports ablations averaged across all benchmarks and backbones. Standard GRPO provides an atomic-only baseline with its own SFT initialization. In the augmented action space, GRPO + Skills improves F1, TSR, and decision compression over the reported SFT baseline. CIPO further improves all three metrics, supporting the benefit of counterfactual granularity training beyond access to skills. Removing the budget constraint from skill mining weakens performance relative to full CIPO, highlighting the importance of a compact and controlled skill library.

\begin{table}[htbp]
\centering
\caption{
Ablation results averaged over all benchmarks and backbones.
$\Delta$F1 and $\Delta$TSR are computed relative to the reported skill-enabled SFT baseline.
C/R denotes Comp./Raw, and Decision red. denotes \(1-\mathrm{C/R}\).
}
\label{tab:ablation}

\resizebox{\linewidth}{!}{
\begin{tabular}{llcccccc}
\toprule
Method & Variant role
& F1\(\uparrow\) & \(\Delta\)F1
& TSR\(\uparrow\) & \(\Delta\)TSR
& C/R\(\downarrow\) & Decision red.\(\uparrow\) \\
\midrule

SFT
& skill-enabled imitation
& 36.05 & 0.00
& 22.91 & 0.00
& 0.958 & 4.2\% \\

Standard GRPO & atomic-only RL & 38.92 & +2.87 & 26.74 & +3.83 & 1.000 & 0.0\% \\

GRPO + Skills
& skill-enabled RL
& 41.20 & +5.15
& 29.30 & +6.39
& 0.912 & 8.8\% \\

w/o Budgeted Mining
& removes mining budget
& 42.82 & +6.77
& 31.75 & +8.84
& 0.872 & 12.8\% \\

CIPO-random
& random eligible branch
& 46.72 & +10.67
& 34.90 & +11.99
& 0.858 & 14.2\% \\

\midrule
\rowcolor{green!8}
\textbf{CIPO-first}
& \textbf{first eligible branch}
& \textbf{48.23} & \textbf{+12.18}
& \textbf{36.48} & \textbf{+13.57}
& \textbf{0.845} & \textbf{15.5\%} \\

\bottomrule
\end{tabular}
}
\end{table}

We further compare CIPO-first with CIPO-random under the same rollout budget and GRPO group construction, where each instruction uses \(G=4\) rollouts with two base rollouts and up to two counterfactual branches. CIPO-random samples a branch point from all eligible granularity decisions in each base rollout, while CIPO-first branches at the first eligible decision. CIPO-first gives the best average F1, TSR, and decision reduction, suggesting that early granularity decisions provide more compute-efficient counterfactual signal. Full ablations, including budgeted mining, executable instantiation, broader branching strategies, and reward sensitivity, are reported in Appendix~\ref{app:full_ablation}.

\section{Conclusion}


We presented CIPO, a counterfactual imagination policy optimization framework for adaptive tool granularity selection in LLM agents. CIPO combines budget-constrained construction of executable skills from successful tool-use trajectories with counterfactual branch rollouts, using paired outcome differences from the same decision context to guide the choice between atomic tools and composite skills. Across three benchmarks and two backbones, CIPO improves tool trajectory quality, task success, and decision compression. These results highlight the importance of learning when to invoke composite skills and when to retain atomic control for effective and efficient tool use.

\clearpage
\bibliography{main}

\appendix

\clearpage

\section{Details of Budget-constrained Skill Mining}
\label{app:mining_details}

This appendix provides additional details for the budget-constrained skill mining procedure described in Section~\ref{sec:method3.2}. The goal of this stage is not to enumerate all frequent subsequences. Instead, the algorithm builds a compact set of recurring tool-use patterns under explicit constraints, which keeps the augmented action space manageable.

\subsection{Input Corpus and Token Abstraction}

The mining stage uses successful rollouts as its input corpus. For each successful rollout, we extract its atomic tool-call sequence and convert each tool invocation into an abstract token. A tool invocation with tool name \(u_t\) and argument dictionary \(p_t\) is represented as
\begin{equation}
z_t = u_t[\mathrm{keys}(p_t)],
\end{equation}
where \(\mathrm{keys}(p_t)\) denotes the sorted set of argument keys. This abstraction removes task-specific argument values while preserving the functional structure of the call. For example, two calls to the same file-reading tool with different file paths are mapped to the same token if both calls require the same argument key. This design reduces spurious variants caused by concrete parameter values and makes repeated tool structures easier to identify.

Let \(\mathcal{D}_{\mathrm{succ}}=\{\mathbf{z}^{(i)}\}_{i=1}^{M}\) denote the resulting corpus of token sequences, where each \(\mathbf{z}^{(i)}=(z^{(i)}_1,\ldots,z^{(i)}_{T_i})\) is extracted from one successful rollout. Sequences with fewer than two tool invocations are ignored because they cannot produce a composite pattern.

\subsection{BPE-style Iterative Merging}

The mining procedure follows the idea of byte-pair encoding, which repeatedly merges frequent adjacent symbols. In our setting, the symbols are tool-call tokens rather than text tokens. At each merge round, the algorithm counts adjacent token pairs over the current corpus and selects the most frequent valid pair:
\begin{equation}
(a^\star,b^\star)=\arg\max_{(a,b)}
\sum_{\mathbf{z}\in\mathcal{D}_{\mathrm{succ}}}
\sum_{t=1}^{|\mathbf{z}|-1}
\mathbb{I}[z_t=a,z_{t+1}=b].
\label{eq:app_bpe_pair_selection}
\end{equation}
The selected pair \((a^\star,b^\star)\) is replaced by a new composite token \(a^\star\oplus b^\star\) at all of its occurrences. The corpus is then updated, and the next round is performed on the updated token sequences. Because composite tokens can be merged again in later rounds, the procedure can discover tool patterns longer than two calls while still relying only on adjacent pair statistics at each step.

The merge history records the selected pair, the new composite token, the pair frequency, the merge round, the resulting atomic length, and the compression gain. The atomic length of a composite token is computed recursively from its merge history. This allows the algorithm to enforce a maximum pattern length even when a candidate is built through several merge rounds.

\subsection{Budget Constraints}

The procedure is controlled by three main constraints. First, the maximum number of merge rounds is \(K\). Second, a selected pair must appear at least \(f_{\min}\) times in the current corpus. Third, the atomic length of the resulting composite token must not exceed \(L_{\max}\). If the most frequent pair violates the length constraint, the algorithm searches the remaining pairs in descending frequency order and selects the highest-frequency pair that satisfies the same constraints. If no valid pair remains, the merge process stops.

After merging, the algorithm applies a pruning step. Each composite token is counted in the final updated corpus. A candidate is removed if its usage is too low, measured by its current frequency and usage ratio, or if its decomposed atomic sequence is shorter than the minimum macro length. This pruning step removes candidates that were useful during intermediate compression but no longer occur often enough after later merges.

In our experiments, we use \(K=50\), \(f_{\min}=3\), and \(L_{\max}=6\) unless otherwise stated. We also use a minimum macro length of 2 and a usage-based pruning threshold. These values were chosen to keep the skill library compact while preserving frequent local routines. We list the implementation values used in our experiments in Table~\ref{tab:mining_hyperparameters}.

\begin{table}[t]
\centering
\small
\caption{Default hyperparameters for budget-constrained skill mining.}
\label{tab:mining_hyperparameters}
\begin{tabular}{lcc}
\toprule
Hyperparameter & Meaning & Default value \\
\midrule
\(K\) & maximum merge rounds & 50 \\
\(f_{\min}\) & minimum adjacent-pair frequency & 3 \\
\(L_{\max}\) & maximum atomic length of a candidate & 6 \\
\(L_{\min}\) & minimum atomic length of a candidate & 2 \\
\(\rho_{\min}\) & minimum usage ratio for pruning & 0.01 \\
\bottomrule
\end{tabular}
\end{table}

\subsection{Size and Length Bounds}

We now state the simple bounds induced by the budget constraints. These bounds are not intended as an optimality guarantee for BPE. They only formalize why the mining stage cannot grow without control.

\paragraph{Proposition 1.}
Let \(K\) be the maximum number of merge rounds. The number of composite tokens generated by the mining procedure is at most \(K\). After pruning, the number of retained symbolic skill candidates is also at most \(K\).

\paragraph{Proof.}
Each merge round creates at most one new composite token. Since the procedure performs no more than \(K\) merge rounds, it can generate at most \(K\) composite tokens before pruning. The pruning stage only removes candidates and never creates new ones. Therefore, the number of retained candidates is no larger than the number of generated composite tokens, which is at most \(K\).

\paragraph{Proposition 2.}
Let \(L_{\max}\) be the maximum allowed atomic length of a composite token. Every retained skill candidate has atomic length at most \(L_{\max}\).

\paragraph{Proof.}
Before a pair is merged, the algorithm computes the atomic length of the resulting composite token. A merge is accepted only if this length is no larger than \(L_{\max}\). Since every generated composite token is produced by an accepted merge, every generated candidate satisfies the length constraint. The pruning stage only removes candidates, so every retained candidate also satisfies the same bound.

These two propositions explain the role of the budget constraints. The number of candidate skills is bounded by \(K\), while the internal length of each candidate is bounded by \(L_{\max}\). As a result, the skill library cannot grow with all frequent subsequences in the trajectory corpus.

\subsection{Computational Complexity}

Let \(N=\sum_{i=1}^{M}|\mathbf{z}^{(i)}|\) be the total number of tokens in the initial corpus. At each round, the algorithm scans the current corpus to count adjacent pairs and then scans it again to replace the selected pair. Since merging only shortens the current corpus, each round costs \(O(N)\) under a straightforward implementation. With at most \(K\) rounds, the overall mining complexity is \(O(KN)\), excluding minor overhead from candidate validation and semantic naming. The memory cost is \(O(N+K)\), where \(O(N)\) stores the current corpus and \(O(K)\) stores the merge history.

The length check can be implemented by caching the atomic length of each token when it is created. With this cache, validating a candidate merge takes constant time after the pair is selected. Our implementation computes lengths from the merge history, which is sufficient for the scale of our experiments. A cached implementation would have the same selected candidates and a lower constant factor.

\subsection{Semantic Skill Naming and Output}

Each retained candidate is decomposed into its atomic tool sequence. The tool names are then used to construct a readable skill identifier. For example, a candidate containing a directory listing tool followed by a file reading tool can be named as a list-and-read style skill. The output of the mining stage stores the skill identifier, merged token, atomic token sequence, tool-name sequence, original frequency, current frequency, usage ratio, merge round, and compression gain.

The output of the mining stage is a symbolic set of reusable tool-use structures. To register these structures as skills, we first decompose each retained composite token into its atomic tool sequence using the merge history. We then assign a skill identifier, attach the corresponding tool schemas, and store the frequency statistics and merge metadata. The registered skill record serves as the input to skill instantiation, where the symbolic sequence is converted into a callable tool with explicit parameters, internal dataflow, selection control, execution traces, and soft interruption.

\section{Details of Executable Skill Instantiation}
\label{app:skill_instantiation}

This appendix describes how a symbolic skill candidate is converted into a callable tool. The mining stage only returns an ordered sequence of atomic tools. Skill instantiation adds a tool schema, parameter interface, execution plan, runtime dataflow, selection control, execution traces, and soft interruption. This conversion is necessary because a frequent tool sequence is not automatically executable under a new task state.

\paragraph{Skill registration.}
For each retained symbolic candidate \(s=(u_1,\ldots,u_m)\), we register a new skill tool and add it to the augmented tool library together with the original atomic tools. Each registered skill contains the following fields:
\begin{itemize}
    \item \texttt{skill\_id}, \texttt{name}, and \texttt{description}: the identifier and natural-language description of the skill;
    \item \texttt{tool\_chain}: the ordered sequence of internal atomic tools;
    \item \texttt{parameters\_schema}: the arguments exposed to the policy;
    \item \texttt{execution\_plan}: the execution order and argument source for each internal tool;
    \item \texttt{pipe\_map}: the mapping from earlier outputs to later inputs;
    \item \texttt{trace\_enabled}: whether internal execution traces are recorded.
\end{itemize}
The registered record is used by the skill interpreter to execute the symbolic sequence as a single callable action.

\paragraph{Parameter exposure and internal piping.}
A symbolic tool sequence usually contains two types of parameters. Some parameters must be provided by the policy because they depend on the current instruction or state. Other parameters can be derived from earlier tool outputs. We separate these cases during registration:
\begin{equation}
\begin{aligned}
P_{\mathrm{exp}}(s) &= \{p \mid p \text{ is supplied by the policy}\},\\
P_{\mathrm{int}}(s) &= \{p \mid p \text{ is resolved from previous outputs}\}.
\end{aligned}
\end{equation}
For the first atomic tool \(u_1\), required parameters are exposed to the policy by default. For each later tool \(u_i\), a parameter is treated as internal if it can be matched to an output field from a previous step. Otherwise, it remains exposed to the policy. This rule keeps the external skill call compact while preserving the dataflow needed for multi-step execution.

\paragraph{Execution templates.}
Each skill is assigned an execution template. The default template is sequential. Under this template, internal tools are executed in order, and later arguments are either supplied by the policy or piped from previous outputs. Some skills require a selection step. This case appears when an earlier tool returns multiple candidates and a later tool requires one target item, identifier, path, or index. For these candidates, the skill exposes a \texttt{select\_strategy} parameter. In our implementation, \texttt{select\_strategy} supports strategies such as \texttt{rank-0}, \texttt{rank-1}, \texttt{random}, and \texttt{filter}. During execution, if an intermediate output is a list, the strategy determines which item is selected before extracting the required field for the next tool.

\paragraph{Runtime execution and soft interruption.}
At runtime, the policy invokes a skill by producing a single tool invocation with the skill name and exposed parameters. The interpreter then executes the internal plan. For each internal step, it resolves arguments from exposed parameters and piped values from earlier outputs. The interpreter records the tool name, arguments, returned observation, and execution status for every internal step.

If an internal step cannot continue, the skill does not fail as an opaque black box. It returns the partial trace, the intermediate outputs that have already been obtained, the interruption step, and the interruption reason. Control is then returned to the policy. The policy may inspect the partial result and continue with atomic tools. For example, if a directory-listing step returns an empty file list, the following file-reading step cannot choose a valid file path. The skill then stops after the listing step, records that no candidate file was available, and returns this partial trace to the policy. If the file list is not empty but the selected file cannot be read, the skill stops at the reading step and returns the selected path together with the failure status. In both cases, the policy can continue from the returned state rather than losing all intermediate information.

\paragraph{Example 1: list-and-read skill.}
Consider a mined candidate
\begin{equation}
s_{\mathrm{read}} =
(\texttt{list\_directory}, \texttt{read\_file}).
\end{equation}
A fixed pipeline would not know which file should be read after the directory is listed. The registered skill therefore exposes the directory path and a selection strategy:
\begin{equation}
\texttt{skill\_list\_and\_read(path, select\_strategy)}.
\end{equation}
Its execution plan is:
\begin{equation}
\begin{aligned}
&1.\quad \texttt{list\_directory(path)} \rightarrow \texttt{files},\\
&2.\quad \texttt{read\_file(path = select(files, select\_strategy))}.
\end{aligned}
\end{equation}
Here, the first \texttt{path} is supplied by the policy. The second \texttt{path} is not supplied directly. It is selected from the file list returned by the first step. If no file can be selected, the skill stops before calling \texttt{read\_file} and returns the partial trace.

\paragraph{Example 2: metadata-to-content skill.}
Consider another mined candidate
\begin{equation}
s_{\mathrm{pdf}} =
(\texttt{get\_pdf\_info}, \texttt{read\_pdf\_pages}).
\end{equation}
This sequence first inspects a PDF and then reads its content. The registered skill can expose the PDF path and the page range:
\begin{equation}
\texttt{skill\_inspect\_and\_read\_pdf(pdf\_path, page\_range)}.
\end{equation}
Its execution plan is:
\begin{equation}
\begin{aligned}
&1.\quad \texttt{get\_pdf\_info(pdf\_path)} \rightarrow \texttt{metadata},\\
&2.\quad \texttt{read\_pdf\_pages(pdf\_path, page\_range)}.
\end{aligned}
\end{equation}
In this example, the policy must provide \texttt{pdf\_path}, because the skill needs to know which document to inspect. The later call to \texttt{read\_pdf\_pages} also requires a page range. If the relevant pages can be inferred from the metadata returned by the first tool, \texttt{page\_range} can be filled internally through the pipe map. Otherwise, it must be exposed as a policy-controlled parameter. In our example, the metadata only describes the document structure and does not determine which pages are relevant to the current task. Therefore, \texttt{page\_range} remains an exposed parameter. This example shows that skill instantiation does not simply pass the first tool's arguments to all later tools. Instead, it specifies which parameters are supplied by the policy and which parameters can be resolved from intermediate outputs.

Skill instantiation turns a symbolic subsequence into an executable action. It exposes parameters that require policy control, pipes parameters that can be resolved from intermediate outputs, supports selection when an internal step returns multiple candidates, and records execution traces for recovery. This design allows composite skills to reduce repeated decisions without becoming fixed and opaque pipelines.

\section{CIPO Training Algorithm}
\label{app:cipo_algorithm}

Algorithm~\ref{alg:cipo} summarizes the training procedure of CIPO. For each instruction, CIPO keeps the total rollout budget fixed at \(G=4\), which matches the GRPO + Skills baseline. It first samples two base rollouts, then attempts to construct at most one counterfactual branch for each base rollout. If fewer than two branches can be constructed, CIPO samples additional base rollouts until the group contains four rollouts. Only valid base-branch pairs receive the granularity-specific supplementary reward. All four rollouts are included in the group-relative policy update.

\begin{algorithm}[H]
\caption{CIPO Training with Fixed Rollout Budget}
\label{alg:cipo}
\begin{algorithmic}[1]
\Require Policy \(\pi_\theta\), old policy \(\pi_{\theta_{\mathrm{old}}}\), augmented action space \(\mathcal{A}\), instruction \(x\), rollout budget \(G=4\), base rollout count \(G_{\mathrm{base}}=2\), base reward \(R_0\)
\State Sample two base rollouts \(\mathcal{B}=\{\tau_{\mathrm{base}}^{(1)},\tau_{\mathrm{base}}^{(2)}\}\) from \(\pi_\theta\)
\State Initialize rollout group \(\mathcal{G}\leftarrow\mathcal{B}\) and pair set \(\mathcal{P}\leftarrow\emptyset\)
\For{each base rollout \(\tau_{\mathrm{base}}\in\mathcal{B}\)}
    \State Scan the actions in \(\tau_{\mathrm{base}}\) from the beginning
    \State Find the first state \(h_t\) with a feasible alternative action \(\tilde{a}_t\) at the other granularity
    \If{such \(h_t\) and \(\tilde{a}_t\) exist}
        \State Fork from the prefix before \(a_t\)
        \State Execute \(\tilde{a}_t\), then continue the interaction with \(\pi_\theta\)
        \State Obtain counterfactual branch \(\tau_{\mathrm{cf}}\)
        \State Add \(\tau_{\mathrm{cf}}\) to \(\mathcal{G}\)
        \State Add \((\tau_{\mathrm{base}},\tau_{\mathrm{cf}})\) to \(\mathcal{P}\)
    \EndIf
\EndFor
\While{\(|\mathcal{G}|<G\)}
    \State Sample an additional base rollout \(\tau_{\mathrm{extra}}\) from \(\pi_\theta\)
    \State Add \(\tau_{\mathrm{extra}}\) to \(\mathcal{G}\)
\EndWhile
\For{each rollout \(\tau_i\in\mathcal{G}\)}
    \State Compute base reward \(R_0(\tau_i)\)
    \State Initialize adjusted reward \(\hat{R}(\tau_i)\leftarrow R_0(\tau_i)\)
\EndFor
\For{each valid pair \((\tau_{\mathrm{base}},\tau_{\mathrm{cf}})\in\mathcal{P}\)}
    \State \(\Delta\leftarrow R_0(\tau_{\mathrm{cf}})-R_0(\tau_{\mathrm{base}})\)
    \State \(r_{\mathrm{cf}}\leftarrow \mathrm{clip}(\alpha\Delta,-\epsilon_{\mathrm{cf}},\epsilon_{\mathrm{cf}})\)
    \State \(\hat{R}(\tau_{\mathrm{cf}})\leftarrow \hat{R}(\tau_{\mathrm{cf}})+r_{\mathrm{cf}}\)
    \State \(\hat{R}(\tau_{\mathrm{base}})\leftarrow \hat{R}(\tau_{\mathrm{base}})-r_{\mathrm{cf}}\)
\EndFor
\State Clip adjusted rewards to be non-negative
\State Normalize \(\{\hat{R}(\tau_i)\}_{i=1}^{G}\) within the group to obtain advantages \(\{\hat{A}_i\}_{i=1}^{G}\)
\State Update \(\pi_\theta\) with the standard GRPO objective
\end{algorithmic}
\end{algorithm}

\section{Counterfactual Replacement Rules}
\label{app:cf_replacement}

This appendix describes how CIPO constructs the alternative action used in a counterfactual branch. The branch does not evaluate a tool output in isolation. Instead, it changes the action granularity at one decision state, executes the alternative action from the same prefix, and lets the same policy complete the remaining interaction.

\paragraph{Scanning rule.}
For each base rollout, CIPO scans the action sequence from the beginning. It creates a branch at the first action where a feasible alternative at the other granularity exists. The prefix before this action is preserved. The alternative action is executed from the same history, and the policy continues from the modified observation.

\paragraph{Skill-to-atomic replacement.}
If the chosen action is a composite skill \(s=(u_1,\ldots,u_m)\), the counterfactual action is the first internal atomic tool \(u_1\). The arguments for \(u_1\) are obtained from the original skill call through the execution plan. In particular, exposed skill parameters are mapped to the required arguments of the first internal tool whenever the mapping is defined. If the required arguments cannot be resolved, no branch is constructed at this action.

\paragraph{Atomic-to-skill replacement.}
If the chosen action is an atomic tool \(a\), CIPO searches the skill library for a skill whose internal tool chain contains \(a\). The candidate set is
\begin{equation}
\mathcal{C}(a)=
\{s\in\mathcal{A}_{\mathrm{skill}} \mid a\in \mathrm{chain}(s)\}.
\end{equation}
A candidate is feasible if its exposed parameters can be constructed from the original atomic arguments and the current history. When multiple candidates are feasible, CIPO selects the shortest feasible skill, with frequency statistics used as a tie breaker. If no feasible skill exists, no counterfactual branch is created at this action.

\paragraph{Argument mapping.}
The replacement action must be meaningful under the same decision context. For a skill-to-atomic replacement, CIPO maps the skill arguments to the schema of the first internal atomic tool. For an atomic-to-skill replacement, CIPO maps the original atomic arguments to the exposed parameters of the selected skill. Parameters that cannot be copied directly may be resolved from the current history or from the skill execution plan. If the mapping fails, the candidate is rejected. This rule prevents CIPO from creating branches with empty or irrelevant arguments.

\begin{cfexample}{Example A: Atomic PDF reading to PDF inspection skill}
\textbf{Base action.}
The base rollout calls an atomic tool:
\[
\texttt{read\_pdf\_pages(pdf\_file\_path, start\_page, end\_page)}.
\]

\textbf{Candidate skill.}
The skill library contains:
\[
\texttt{skill\_get\_pdf\_and\_read\_pdf\_pages}
\]
with internal chain:
\[
\texttt{get\_pdf\_info}
\rightarrow
\texttt{read\_pdf\_pages}.
\]

\textbf{Replacement.}
The skill chain contains \(\texttt{read\_pdf\_pages}\), so CIPO can replace the atomic call with this skill if the exposed parameters can be filled. The original atomic arguments provide \(\texttt{pdf\_file\_path}\), \(\texttt{start\_page}\), and \(\texttt{end\_page}\), which are sufficient for the skill call.

\textbf{Counterfactual branch.}
The branch first executes \(\texttt{get\_pdf\_info}\), then executes \(\texttt{read\_pdf\_pages}\) with the same page range. This branch tests whether adding metadata inspection before reading improves the later trajectory.
\end{cfexample}

\begin{cfexample}{Example B: PDF inspection skill to first atomic tool}
\textbf{Base action.}
The base rollout calls:
\[
\texttt{skill\_get\_pdf\_and\_read\_pdf\_pages(pdf\_file\_path,start\_page,end\_page)}.
\]

\textbf{Internal chain.}
The skill executes:
\[
\texttt{get\_pdf\_info}
\rightarrow
\texttt{read\_pdf\_pages}.
\]

\textbf{Replacement.}
For a skill-to-atomic branch, CIPO uses the first internal atomic tool:
\[
\tilde{a}=\texttt{get\_pdf\_info(pdf\_file\_path)}.
\]
The argument \(\texttt{pdf\_file\_path}\) is copied from the original skill call through the execution plan.

\textbf{Counterfactual branch.}
The branch only performs metadata inspection at the branch point. The policy then decides whether to read pages, call another tool, or revise its plan. This branch tests whether preserving control after the first atomic step is better than executing the full skill immediately.
\end{cfexample}

\begin{cfexample}{Example C: Atomic file reading to list-and-read skill}
\textbf{Base action.}
The base rollout calls:
\[
\texttt{read\_file(path="/workspace/report.txt")}.
\]

\textbf{Candidate skill.}
The skill library contains:
\[
\texttt{skill\_list\_and\_read(path, select\_strategy)}
\]
with internal chain:
\[
\texttt{list\_directory}
\rightarrow
\texttt{read\_file}.
\]

\textbf{Feasibility check.}
The skill contains \(\texttt{read\_file}\), but it also requires a directory path and a selection strategy. If the directory can be derived from the original file path, CIPO maps \(\texttt{/workspace/report.txt}\) to \(\texttt{/workspace}\). The selection strategy can be filled by a default strategy or by a policy-controlled parameter.

\textbf{Counterfactual branch.}
When the mapping is feasible, the branch lists the directory, selects a file according to \(\texttt{select\_strategy}\), and then reads it. This branch tests whether a broader list-and-read routine helps, or whether the original direct file read was more appropriate.
\end{cfexample}

\begin{cfexample}{Example D: Write-and-run skill to atomic file writing}
\textbf{Base action.}
The base rollout calls:
\[
\texttt{skill\_write\_file\_and\_run\_cmd(content,path,cmd)}.
\]

\textbf{Internal chain.}
The skill executes:
\[
\texttt{write\_file}
\rightarrow
\texttt{run\_cmd}.
\]

\textbf{Replacement.}
The counterfactual action is the first internal atomic tool:
\[
\tilde{a}=\texttt{write\_file(content,path)}.
\]

\textbf{Counterfactual branch.}
After writing the file, the policy regains control. It may run the command, inspect the file, modify the script, or use another tool. This branch tests whether executing the full write-and-run routine is better than preserving control after file creation.
\end{cfexample}

\begin{cfexample}{Example E: Infeasible atomic-to-skill replacement}
\textbf{Base action.}
The base rollout calls:
\[
\texttt{update\_cells(spreadsheet\_id, range, values)}.
\]

\textbf{Candidate skill.}
A candidate skill contains:
\[
\texttt{read\_sheet}
\rightarrow
\texttt{update\_cells}.
\]

\textbf{Feasibility check.}
Although the skill chain contains \(\texttt{update\_cells}\), the skill may require a sheet name, a read range, or a selection rule that cannot be recovered from the original update call or the current history. In this case, the replacement would not be executable from the same decision context.

\textbf{Decision.}
CIPO rejects this candidate and continues scanning later actions in the base rollout. This example shows that CIPO does not replace every atomic tool with any skill that contains it. The alternative action must be executable with meaningful arguments.
\end{cfexample}

\section{Reward Composition}
\label{app:reward_composition}

This appendix describes the reward used during CIPO training. The reward has two layers. First, each rollout receives a base reward \(R_0\) from task progress, skill behavior, and execution efficiency. Second, valid base-branch pairs receive a bounded counterfactual adjustment that provides feedback for tool granularity selection.

\paragraph{Base reward.}
Before counterfactual adjustment, each rollout receives
\begin{equation}
R_0(\tau)
=
R_{\mathrm{task}}(\tau)
+
R_{\mathrm{skill}}(\tau)
+
R_{\mathrm{eff}}(\tau).
\end{equation}
The task component measures whether the rollout covers the reference tools. The skill component rewards useful skill calls and penalizes irrelevant or repeated skill calls. The efficiency component rewards shorter decision trajectories only when the rollout has sufficient task progress.

\paragraph{Task component.}
\(R_{\mathrm{task}}\) is computed from coverage of the reference tool sequence. Tool names are normalized before matching. Exact matches receive full credit, substring matches receive partial credit, and token-overlap matches receive credit when their Jaccard score is sufficiently high. The coverage score is combined with an order-preservation bonus, so a rollout receives higher task reward when it covers the correct tools in a more consistent order. If the rollout produces no tool invocations, it receives a zero-step penalty.

\paragraph{Skill component.}
\(R_{\mathrm{skill}}\) is an additive term. A skill receives positive credit only when its internal tool chain overlaps with the reference tools. A fully successful relevant skill receives the largest bonus, while a partially successful relevant skill receives a smaller bonus according to the fraction of successful internal steps. If a skill has no overlap with the reference tools, it receives a penalty. Repeated calls to the same skill are also penalized. The total skill bonus is clipped to avoid a collapse in which the policy calls skills only to obtain reward.

\paragraph{Efficiency component.}
\(R_{\mathrm{eff}}\) encourages shorter decision trajectories, but only when the rollout has sufficient task progress. In our implementation, the efficiency bonus is gated by task completion and task coverage. A compression bonus is added when the rollout uses skills to reduce the number of policy decisions. Very short rollouts with low task coverage receive an under-exploration penalty. This design prevents the policy from receiving high reward for stopping early without solving the task.

\paragraph{Counterfactual adjustment.}
For each valid base--branch pair, CIPO computes the base-reward difference
\begin{equation}
\Delta
=
R_0(\tau_{\mathrm{cf}})
-
R_0(\tau_{\mathrm{base}}).
\end{equation}
The difference is converted into a bounded supplementary reward:
\begin{equation}
r_{\mathrm{cf}}
=
\operatorname{clip}
\left(
\alpha \Delta,
-\epsilon_{\mathrm{cf}},
\epsilon_{\mathrm{cf}}
\right).
\end{equation}
CIPO applies this signal symmetrically to the paired rollouts:
\begin{equation}
\hat{R}(\tau_{\mathrm{cf}})
=
R_0(\tau_{\mathrm{cf}})+r_{\mathrm{cf}},
\qquad
\hat{R}(\tau_{\mathrm{base}})
=
R_0(\tau_{\mathrm{base}})-r_{\mathrm{cf}}.
\end{equation}
Rollouts without a valid counterfactual pair retain their base reward before clipping.
After all paired adjustments, we apply
\(\hat{R}(\tau_i)\leftarrow\max\{0,\hat{R}(\tau_i)\}\)
to every rollout before group-relative normalization.

\paragraph{Group-relative normalization.}
For each instruction, CIPO forms a group of \(G=4\) rollouts. The group contains two base rollouts, valid counterfactual branches, and additional base rollouts used to fill the fixed budget when fewer branches are available. The adjusted reward of rollout \(i\) is normalized within the group:
\begin{equation}
\hat{A}_i
=
\frac{
\hat{R}_i-\mathrm{mean}(\{\hat{R}_j\}_{j=1}^{G})
}{
\mathrm{std}(\{\hat{R}_j\}_{j=1}^{G})+\eta
}.
\end{equation}
CIPO then applies the standard GRPO objective. For generated token \(t\) in rollout \(i\), define
\begin{equation}
\rho_{i,t}(\theta)
=
\frac{
\pi_\theta(y_{i,t}\mid h_{i,t})
}{
\pi_{\theta_{\mathrm{old}}}(y_{i,t}\mid h_{i,t})
}.
\end{equation}
The policy objective is
\begin{equation}
\begin{aligned}
\mathcal{J}_{\mathrm{CIPO}}(\theta)
=
\frac{1}{G}\sum_{i=1}^{G}\frac{1}{|y_i|}
\sum_{t=1}^{|y_i|}
\Bigg[
&\min\Big(
\rho_{i,t}(\theta)\hat{A}_i,\,
\mathrm{clip}\big(
\rho_{i,t}(\theta),
1-\epsilon_{\mathrm{clip}},
1+\epsilon_{\mathrm{clip}}
\big)\hat{A}_i
\Big) \\
&-
\beta D_{\mathrm{KL}}\Big[
\pi_\theta(\cdot\mid h_{i,t})
\Vert
\pi_{\mathrm{ref}}(\cdot\mid h_{i,t})
\Big]
\Bigg].
\end{aligned}
\end{equation}
CIPO therefore preserves the standard GRPO optimization form. Its difference lies in how the rollout group is constructed and how the reward is adjusted before group-relative normalization.

\paragraph{Role of the counterfactual term.}
The counterfactual reward is a bounded supplementary signal, not a standalone causal estimate. The branch compares two rollouts that share the same prefix but differ in one granularity choice, after which both continue under the current policy. This comparison gives direct feedback about whether the atomic or skill action worked better under the matched decision context. The task reward remains the main optimization signal throughout training.

\section{Training Dynamics}
\label{app:training_dynamics}

\begin{figure*}[ht]
    \centering
    \includegraphics[width=\textwidth]{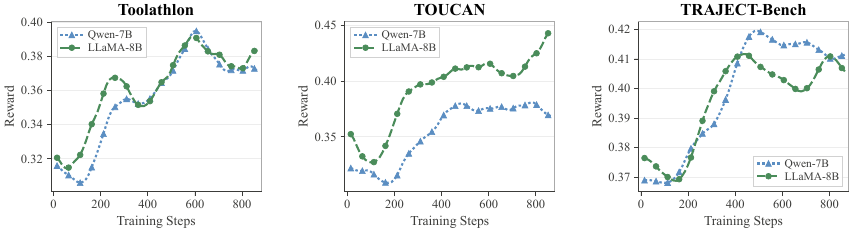}
    \caption{
    Training reward curves of CIPO across three benchmarks and two backbone models. The reward generally increases during training across TOOLATHLON, TOUCAN, and TRAJECT-Bench, suggesting that the counterfactual branch signal can be optimized together with the task reward.
    }
    \label{fig:app_training_dynamics}
\end{figure*}

We provide the training reward curves of CIPO in Figure~\ref{fig:app_training_dynamics}. The curves are reported for three benchmarks, TOOLATHLON, TOUCAN, and TRAJECT-Bench, with Qwen-7B and Llama-8B as backbone models. Here Qwen-7B and Llama-8B denote Qwen2.5-7B and Llama3.1-8B, respectively. The logged reward is the group-level training reward used during policy optimization, which includes task progress, skill behavior, efficiency, and the bounded counterfactual adjustment described in Appendix~\ref{app:reward_composition}. These curves are used to analyze the optimization process. The final evaluation results are reported separately in Table~\ref{tab:main_results}.

Overall, CIPO shows a consistent upward trend across the evaluated settings. On TOOLATHLON, both backbones improve after the early training stage and reach substantially higher rewards than their initial values. The curves contain local fluctuations, but the overall direction remains positive, indicating that the policy gradually learns to use the augmented action space more effectively. On TOUCAN, the improvement is especially clear for Llama-8B, whose reward increases steadily throughout training. Qwen-7B also improves and then enters a relatively stable region, suggesting that the optimization signal remains useful after the policy has learned many common tool-use patterns. On TRAJECT-Bench, both models improve rapidly in the early and middle stages. The reward later fluctuates around a higher plateau, which is expected because this benchmark contains more diverse trajectories and the sampled rollouts may vary in difficulty across updates.

The reward curves are not expected to be strictly monotonic. CIPO samples rollouts online, and each update can contain different task instances, base rollouts, and counterfactual branches. In addition, the counterfactual branch may expose a better or worse granularity choice under the current policy, which introduces controlled variation in the group rewards. These fluctuations are useful during training because they provide contrastive feedback for deciding between atomic tools and composite skills. The important observation is that the rewards increase or stabilize at higher levels across all benchmark-backbone combinations, which supports the stability of the CIPO optimization process.

These dynamics also help interpret the main results. The improvement is not limited to a single benchmark or a single backbone, suggesting that the method does not rely on a benchmark-specific skill library or a model-specific behavior. Instead, CIPO provides a general training signal for adaptive tool granularity selection. The main evaluation in Table~\ref{tab:main_results} further shows that this training trend translates into better Tool F1, TSR, and decision compression.

\section{Derivation of the Coverage-derived Skill Usage Reference}
\label{app:coverage_reference}

This appendix explains how we compute the coverage-derived reference used in Figure~\ref{fig:skill_usage}. This reference is not a learned policy. It is an estimate of the expected skill-call ratio when the atomic steps covered by the mined skill library are compressed into composite skill calls.

Let an atomic-only trajectory contain \(N\) atomic tool invocations. Suppose the mined skill library can cover a fraction \(c\) of these atomic calls. Then \(cN\) atomic calls are replaceable by skills, while the remaining \((1-c)N\) atomic calls are not covered by the skill library and must remain as atomic decisions.

Let \(m\) denote the frequency-weighted average length of the mined skills. If the covered atomic calls are compressed by skills of average length \(m\), then the \(cN\) covered atomic calls are replaced by approximately
\begin{equation}
\frac{cN}{m}
\label{eq:covered_skill_calls}
\end{equation}
skill calls. The uncovered part still requires
\begin{equation}
(1-c)N
\label{eq:uncovered_atomic_calls}
\end{equation}
atomic tool invocations. Therefore, after compression, the expected number of policy-level decisions is
\begin{equation}
\frac{cN}{m} + (1-c)N,
\label{eq:expected_policy_decisions}
\end{equation}
where the number of skill decisions is \(cN/m\). The expected skill usage ratio is therefore
\begin{equation}
\mathrm{SkillRatio}_{\mathrm{ref}}
=
\frac{cN/m}{cN/m+(1-c)N}
=
\frac{c/m}{c/m+(1-c)}.
\label{eq:coverage_reference}
\end{equation}

This quantity reflects the expected fraction of policy-level decisions that would be skill calls if all replaceable atomic steps were compressed by available skills. The coverage \(c\) increases the reference ratio because more atomic steps can be replaced by skills. The average length \(m\) has the opposite effect on the number of skill calls because a longer skill replaces more atomic calls with a single decision. The uncovered fraction \(1-c\) remains as atomic decisions and reduces the expected skill-call ratio.

We use this value as a coverage-derived reference level in Figure~\ref{fig:skill_usage}. It should not be interpreted as a state-level optimal policy. Instead, it provides a library-level reference for judging whether a trained policy uses skills at a plausible rate given the amount of reusable structure available in the skill library.
The reference in Figure~\ref{fig:skill_usage} is computed for the TOOLATHLON skill library used with Qwen2.5-7B and is used only to analyze training-time behavior.
Figure~\ref{fig:skill_usage} measures skill usage in online training rollouts, whereas Table~\ref{tab:additional_metrics} reports skill usage on the final evaluation trajectories.
The two therefore characterize different rollout distributions, and their absolute skill-usage ratios are not expected to match.

\section{Full Ablation Results}
\label{app:full_ablation}

This appendix provides the full ablation results across all benchmark-backbone settings. The main paper reports averaged ablation results for readability. Here, we expand each ablation into six settings: TOOLATHLON with Qwen-7B and Llama-8B, TOUCAN with Qwen-7B and Llama-8B, and TRAJECT-Bench with Qwen-7B and Llama-8B. C/R denotes Comp./Raw, where lower values indicate stronger decision compression.

\begin{table*}[t]
\centering
\scriptsize
\setlength{\tabcolsep}{2.8pt}
\renewcommand{\arraystretch}{1.08}
\caption{
Detailed ablation results across all benchmark-backbone settings. Each entry reports F1 / TSR. Avg. reports the arithmetic mean over the six settings.
}
\label{tab:full_ablation_f1_tsr}
\resizebox{\textwidth}{!}{
\begin{tabular}{lcccccccc}
\toprule
\multirow{2}{*}{Method}
& \multicolumn{2}{c}{TOOLATHLON}
& \multicolumn{2}{c}{TOUCAN}
& \multicolumn{2}{c}{TRAJECT-Bench}
& \multirow{2}{*}{Avg. F1}
& \multirow{2}{*}{Avg. TSR} \\
\cmidrule(lr){2-3}\cmidrule(lr){4-5}\cmidrule(lr){6-7}
& Qwen-7B & Llama-8B
& Qwen-7B & Llama-8B
& Qwen-7B & Llama-8B
&  &  \\
\midrule
SFT
& 25.88 / 15.24 & 39.42 / 23.58
& 34.56 / 19.72 & 36.71 / 22.84
& 37.04 / 24.30 & 42.69 / 31.78
& 36.05 & 22.91 \\

Standard GRPO
& 28.90 / 18.42 & 42.10 / 27.96
& 38.25 / 24.70 & 40.68 / 27.65
& 40.20 / 27.80 & 43.39 / 33.91
& 38.92 & 26.74 \\

GRPO + Skills
& 31.10 / 22.20 & 44.35 / 32.55
& 40.20 / 30.10 & 42.30 / 31.20
& 42.85 / 29.10 & 46.40 / 30.65
& 41.20 & 29.30 \\

w/o Budgeted Mining
& 31.40 / 24.20 & 45.10 / 34.20
& 42.90 / 31.50 & 44.60 / 33.20
& 44.80 / 30.10 & 48.10 / 37.30
& 42.82 & 31.75 \\

w/o Executable Instantiation
& 31.50 / 23.70 & 44.80 / 33.60
& 41.80 / 30.70 & 43.70 / 32.50
& 43.60 / 29.80 & 47.20 / 36.00
& 42.10 & 31.05 \\

w/o CF Reward
& 32.84 / 25.63 & 46.72 / 36.40
& 42.80 / 31.70 & 44.95 / 33.80
& 44.60 / 29.80 & 47.59 / 32.57
& 43.25 & 31.65 \\

CIPO-late
& 33.90 / 28.50 & 47.20 / 37.90
& 45.60 / 33.40 & 47.25 / 38.20
& 48.10 / 30.20 & 52.51 / 36.40
& 45.76 & 34.10 \\

CIPO-random
& 34.28 / 28.96 & 47.65 / 38.82
& 46.20 / 34.00 & 48.65 / 39.35
& 48.61 / 30.72 & 54.93 / 37.55
& 46.72 & 34.90 \\

CIPO-all
& 34.86 / 29.10 & \textbf{49.05} / 39.80
& 47.20 / 33.40 & 49.72 / 39.20
& 48.40 / 30.65 & \textbf{56.39} / 37.85
& 47.60 & 35.00 \\

\midrule
\rowcolor{blue!8}
\textbf{CIPO-first}
& \textbf{35.17} / \textbf{30.78}
& 48.32 / \textbf{40.55}
& \textbf{48.55} / \textbf{35.50}
& \textbf{50.51} / \textbf{41.32}
& \textbf{50.99} / \textbf{32.00}
& 55.86 / \textbf{38.75}
& \textbf{48.23}
& \textbf{36.48} \\
\bottomrule
\end{tabular}
}
\end{table*}

\begin{table*}[t]
\centering
\scriptsize
\setlength{\tabcolsep}{4.2pt}
\renewcommand{\arraystretch}{1.08}
\caption{
Detailed decision compression results for ablation variants. C/R denotes Comp./Raw. Lower values indicate stronger decision compression.
}
\label{tab:full_ablation_cr}
\resizebox{\textwidth}{!}{
\begin{tabular}{lccccccc}
\toprule
\multirow{2}{*}{Method}
& \multicolumn{2}{c}{TOOLATHLON}
& \multicolumn{2}{c}{TOUCAN}
& \multicolumn{2}{c}{TRAJECT-Bench}
& \multirow{2}{*}{Avg. C/R} \\
\cmidrule(lr){2-3}\cmidrule(lr){4-5}\cmidrule(lr){6-7}
& Qwen-7B & Llama-8B
& Qwen-7B & Llama-8B
& Qwen-7B & Llama-8B
&  \\
\midrule
SFT
& 0.960 & 0.940
& 0.950 & 0.930
& 1.000 & 0.970
& 0.958 \\

Standard GRPO & 1.000 & 1.000 & 1.000 & 1.000 & 1.000 & 1.000 & 1.000 \\

GRPO + Skills
& 0.920 & 0.880
& 0.910 & 0.860
& 0.960 & 0.942
& 0.912 \\

w/o Budgeted Mining
& 0.880 & 0.820
& 0.860 & 0.790
& 0.950 & 0.932
& 0.872 \\

w/o Executable Instantiation
& 0.873 & 0.815
& 0.846 & 0.781
& 0.960 & 0.921
& 0.866 \\

w/o CF Reward
& 0.891 & 0.842
& 0.873 & 0.809
& 0.958 & 0.931
& 0.884 \\

CIPO-late
& 0.875 & 0.805
& 0.870 & 0.835
& 0.935 & 0.846
& 0.861 \\

CIPO-random
& 0.906 & 0.800
& 0.860 & 0.825
& 0.925 & 0.832
& 0.858 \\

CIPO-all
& \textbf{0.830} & \textbf{0.765}
& \textbf{0.820} & \textbf{0.775}
& 0.960 & 0.950
& 0.850 \\

\midrule
\rowcolor{blue!8}
\textbf{CIPO-first}
& 0.855 & 0.772
& 0.860 & 0.812
& \textbf{0.910} & \textbf{0.861}
& \textbf{0.845} \\
\bottomrule
\end{tabular}
}
\end{table*}

\paragraph{Component effects.}
Standard GRPO provides an atomic-only comparison with a separate SFT initialization. In the augmented action space, GRPO + Skills improves F1, TSR, and decision compression over the reported SFT baseline. CIPO further improves these metrics. Removing the counterfactual reward while retaining branch rollouts weakens performance, supporting the contribution of the paired outcome difference to granularity training.

\paragraph{Skill construction and instantiation.}
Removing budgeted mining reduces performance because the skill library becomes more redundant and harder to select from. Removing executable instantiation also hurts performance, showing that symbolic subsequences are not enough. Skills need explicit parameter interfaces, internal dataflow, selection control, execution traces, and soft interruption to work reliably in new task states.

\paragraph{Branch-point selection.}
CIPO-random and CIPO-late keep the same rollout budget and group construction as CIPO-first, using \(G=4\) rollouts with two base rollouts and at most two counterfactual branches. They differ only in how the branch point is selected. CIPO-random samples from all eligible granularity decisions, while CIPO-late prefers later eligible decisions. Both variants are weaker than CIPO-first on average, suggesting that the first eligible decision gives a cleaner comparison under the fixed group budget.

CIPO-all is an extra-compute variant that branches at every eligible granularity decision in each base rollout. Since it no longer keeps the group size fixed, its rollout cost grows with the number of eligible branch points. It improves some individual settings, such as Llama-8B on TOOLATHLON and TRAJECT-Bench, and achieves stronger compression on TOOLATHLON and TOUCAN. However, its average F1 and TSR remain below CIPO-first, and its compression becomes worse on TRAJECT-Bench. This indicates that broader branching is not automatically better once computation and reward noise are considered, supporting first-eligible branching as a more stable and compute-efficient default.

\begin{figure*}[ht]
    \centering
    \includegraphics[width=\textwidth]{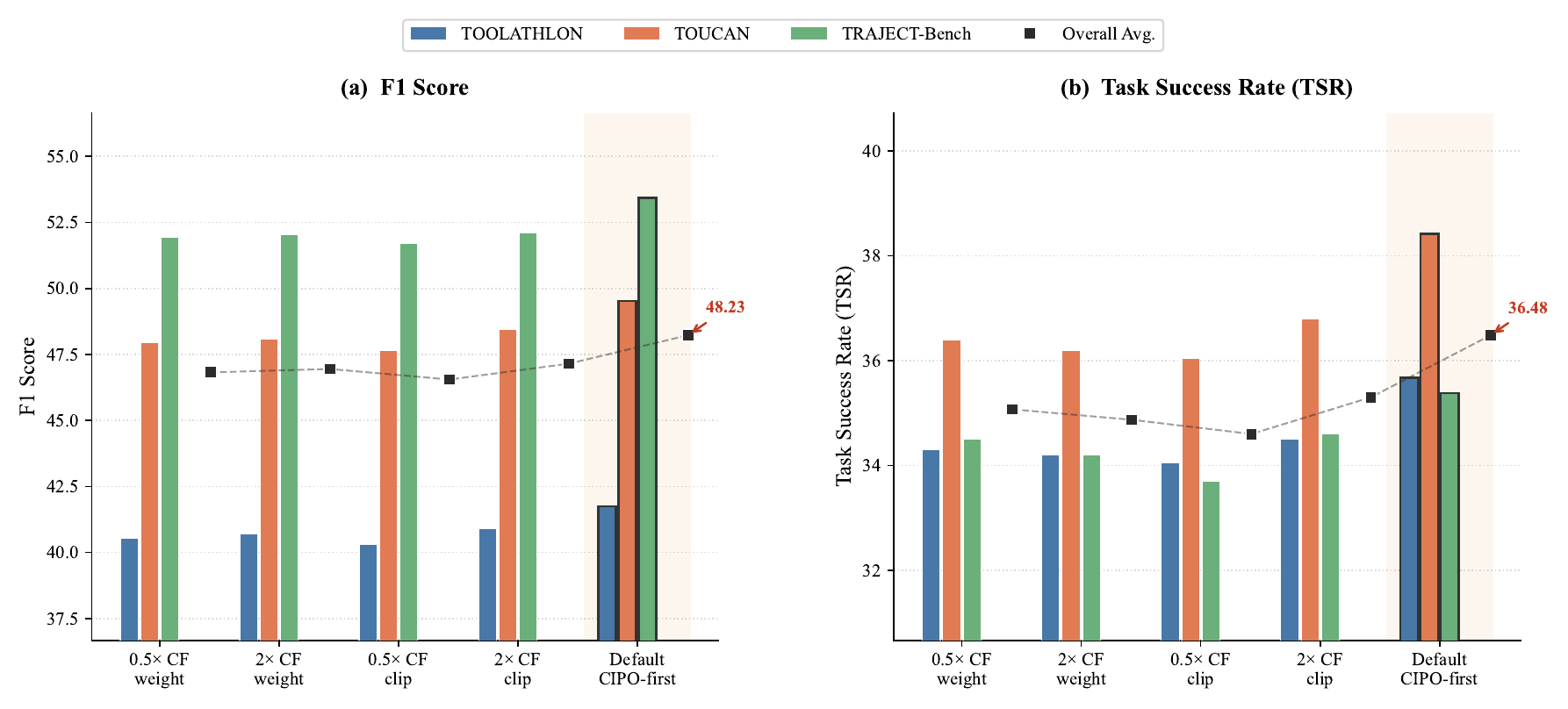}
\caption{
Sensitivity analysis of the counterfactual reward. We vary one hyperparameter at a time around the default setting, including the reward scale \(\alpha\) and the clipping bound \(\epsilon_{\mathrm{cf}}\). Default CIPO-first achieves the best overall balance across benchmarks and models.
}
\label{fig:reward_sensitivity}
\end{figure*}

\paragraph{Reward sensitivity.}
Figure~\ref{fig:reward_sensitivity} evaluates the sensitivity of CIPO to the counterfactual reward scale \(\alpha\) and clipping bound \(\epsilon_{\mathrm{cf}}\). Each variant changes one hyperparameter while keeping the other settings fixed. Reducing \(\alpha\) weakens the granularity signal, while increasing \(\alpha\) can overemphasize paired reward differences. Similarly, a narrow clipping range suppresses useful counterfactual differences, whereas a wide clipping range can make the supplementary signal too dominant. The default setting used in the main experiments achieves the highest overall F1 and TSR among these variants, indicating that CIPO benefits from a bounded counterfactual signal that is strong enough to guide granularity selection but not strong enough to override the task reward.

\section{Hyperparameters and Training Details}
\label{app:hyperparameters}

This appendix reports the hyperparameters and implementation details used in our experiments. Qwen-7B and Llama-8B denote Qwen2.5-7B and Llama3.1-8B, respectively. We omit machine-specific paths and private credentials from this description.

\paragraph{Overall pipeline.}
CIPO follows a five-stage pipeline. First, recurring tool-use patterns are mined from successful trajectories. Second, the retained candidates are instantiated as executable skills and combined with the original atomic tools to form the augmented tool library. Third, the model is supervised fine-tuned on successful trajectories in the augmented action space. Fourth, the policy is optimized with CIPO using counterfactual branch rollouts. Finally, the trained policy is evaluated in the augmented action space without counterfactual branching. The same pipeline is used for both Qwen-7B and Llama-8B.

\paragraph{Skill mining.}
We use budget-constrained BPE-style mining over successful tool-use trajectories. Each tool invocation is abstracted by its tool name and parameter-key signature, which removes task-specific argument values while preserving tool structure. Table~\ref{tab:skill_mining_hparams} lists the mining hyperparameters.

\begin{table}[ht]
\centering
\small
\caption{Hyperparameters for budget-constrained skill mining.}
\label{tab:skill_mining_hparams}
\begin{tabular}{lc}
\toprule
Hyperparameter & Value \\
\midrule
Input trajectories & successful trajectories only \\
Token abstraction & tool name + parameter-key signature \\
Maximum merge rounds \(K\) & 50 \\
Minimum pair frequency \(f_{\min}\) & 3 \\
Minimum skill length & 2 atomic tools \\
Maximum skill length \(L_{\max}\) & 6 atomic tools \\
Minimum usage ratio for pruning & 0.01 \\
\bottomrule
\end{tabular}
\end{table}

\paragraph{Skill instantiation.}
The mined symbolic candidates are converted into executable skills before being exposed to the policy. Each skill is registered with a parameter schema, execution plan, internal dataflow map, execution trace, and interruption behavior. Table~\ref{tab:skill_instantiation_hparams} summarizes the instantiation settings.

\begin{table}[ht]
\centering
\small
\caption{Skill instantiation settings.}
\label{tab:skill_instantiation_hparams}
\begin{tabular}{lc}
\toprule
Item & Value \\
\midrule
Execution templates & sequential, select, conditional \\
Selection strategies & rank-0, rank-1, random, filter \\
Trace recording & enabled \\
Soft interruption & enabled \\
Skill interface & exposed parameters + internal piping \\
\bottomrule
\end{tabular}
\end{table}

\paragraph{Supervised fine-tuning.}
The skill-enabled SFT stage initializes the policy in the augmented action space. It teaches the model the tool-call format and exposes it to both atomic tools and composite skills before policy optimization. We use LoRA fine-tuning and compute the loss on assistant tokens only. The SFT hyperparameters are shown in Table~\ref{tab:sft_hparams}. Standard GRPO uses a separate atomic-only SFT initialization, as detailed in Appendix~\ref{app:grpo_baselines}.

\begin{table}[ht]
\centering
\small
\caption{Skill-enabled supervised fine-tuning hyperparameters.}
\label{tab:sft_hparams}
\begin{tabular}{lc}
\toprule
Hyperparameter & Value \\
\midrule
Training data & successful trajectories in augmented action space \\
Epochs & 3 \\
Learning rate & \(1\times10^{-5}\) \\
Per-device batch size & 2 \\
Gradient accumulation steps & 8 \\
Maximum sequence length & 4096 \\
Warmup ratio & 0.05 \\
LoRA rank & 64 \\
LoRA alpha & 128 \\
LoRA dropout & 0.10 \\
LoRA target modules & q, k, v, o, gate, up, down projections \\
Weight decay & 0.01 \\
Save steps & 200 \\
Logging steps & 10 \\
Loss mask & assistant tokens only \\
\bottomrule
\end{tabular}
\end{table}

\paragraph{CIPO policy optimization.}
CIPO starts from the SFT checkpoint and optimizes the policy with group-relative policy optimization. In the paper implementation, we keep the rollout budget fixed at \(G=4\) for each instruction. Each group contains two base rollouts, up to two counterfactual branches, and additional base rollouts when fewer than two valid branches are available. All four rollouts are used for group-relative normalization, while only valid base-branch pairs receive the counterfactual supplementary reward. Table~\ref{tab:cipo_hparams} gives the policy-optimization hyperparameters used for the 7B and 8B backbones.

\begin{table}[ht]
\centering
\small
\caption{CIPO policy-optimization hyperparameters for Qwen-7B and Llama-8B.}
\label{tab:cipo_hparams}
\begin{tabular}{lc}
\toprule
Hyperparameter & Value \\
\midrule
Initialization & SFT checkpoint \\
Action space & atomic tools + executable skills \\
Epochs & 3 \\
Learning rate & \(2\times10^{-6}\) \\
Rollout group size \(G\) & 4 \\
Base rollouts per instruction & 2 \\
Counterfactual branches per instruction & up to 2 \\
Fallback for missing branches & additional base rollouts \\
Branch point & first eligible granularity decision \\
KL coefficient \(\beta\) & 0.05 \\
Sampling temperature & 0.7 \\
Per-device batch size & 1 \\
Gradient accumulation steps & 4 \\
Maximum sequence length & 3072 \\
Maximum generation length & 384 \\
LoRA rank & 32 \\
LoRA alpha & 64 \\
LoRA dropout & 0.05 \\
LoRA target modules & q, k, v, o, gate, up, down projections \\
Gradient checkpointing & enabled \\
Save steps & 50 \\
Logging steps & 1 \\
\bottomrule
\end{tabular}
\end{table}

\paragraph{Reward configuration.}
The base reward combines task progress, skill behavior, and efficiency terms. The counterfactual branch adds a bounded supplementary signal based on the paired reward difference. Table~\ref{tab:reward_hparams} lists the reward hyperparameters.

\begin{table}[ht]
\centering
\small
\caption{Reward hyperparameters.}
\label{tab:reward_hparams}
\begin{tabular}{lc}
\toprule
Hyperparameter & Value \\
\midrule
Skill reward weight \(\lambda_{\mathrm{skill}}\) & 0.30 \\
Complete skill reward & 1.00 \\
Passed internal-step reward & 0.20 \\
Failed internal-step penalty & \(-0.50\) \\
Efficiency reward gate & 0.40 \\
Under-exploration penalty & 0.05 \\
Under-exploration threshold & 3 \\
Counterfactual reward scale \(\alpha\) & 0.15 \\
Counterfactual clipping bound $\epsilon_{\mathrm{cf}}$ & 0.10 \\
\bottomrule
\end{tabular}
\end{table}

\paragraph{Counterfactual branch construction.}
For each base rollout, CIPO scans the generated actions from the beginning and branches at the first eligible granularity decision. If the chosen action is an atomic tool, the branch replaces it with a feasible skill whose internal chain contains that tool. If the chosen action is a skill, the branch replaces it with the first internal atomic tool of the skill. The branch starts from the same prefix, executes the replacement action, and lets the current policy complete the remaining interaction. This keeps the comparison tied to the same decision context while preserving a fixed group budget.

\paragraph{Large-model training setup.}
The 7B and 8B experiments use a two-GPU training path. The model is loaded with automatic device mapping, gradient checkpointing is enabled, and LoRA is used to keep the number of trainable parameters small. Rollout generation is performed online, and each rollout group is immediately used for reward computation and policy optimization.

\paragraph{Evaluation protocol.}
At inference time, CIPO directly selects actions from the augmented tool library without counterfactual branching. Evaluation uses deterministic decoding and a fixed execution budget. Table~\ref{tab:eval_hparams} lists the evaluation settings for CIPO.

\begin{table}[ht]
\centering
\small
\caption{CIPO evaluation hyperparameters.}
\label{tab:eval_hparams}
\begin{tabular}{lc}
\toprule
Hyperparameter & Value \\
\midrule
Evaluation episodes per setting & 100 \\
Maximum turns & 30 \\
Maximum expanded atomic calls & 50 \\
Inference temperature & 0.0 \\
Top-p & 0.95 \\
Batch size & 1 \\
Continuation mechanism & enabled \\
Inference action space & atomic tools + executable skills \\
Counterfactual branches at inference & disabled \\
Main reported metrics & Tool F1, TSR, Comp./Raw \\
\bottomrule
\end{tabular}
\end{table}

\paragraph{Metric computation.}
Tool F1 is computed as an order-aware match between the predicted tool-use trajectory and the reference trajectory after tool-name normalization. TSR is evaluated by an LLM judge, as described in Appendix~\ref{app:llm_judge}. Comp./Raw compares policy-level decisions after skill compression with the expanded atomic calls of the same executed trajectory. Additional metrics and confidence intervals are provided in Appendix~\ref{app:additional_results}.

\section{Additional Results and Uncertainty Estimates}
\label{app:additional_results}

\paragraph{Additional evaluation metrics.}
Table~\ref{tab:additional_metrics} reports additional CIPO metrics that are not shown in the main table. We include trajectory-quality metrics, skill-use metrics, reliability metrics, and decision-efficiency details. Prec. and Rec. denote tool precision and recall. NTAcc denotes next-tool accuracy, which measures whether the next selected tool matches the reference trajectory at each step. Prog. denotes progress rate, indicating the fraction of episodes with non-zero task progress. Skill Use denotes the fraction of policy-level decisions that invoke composite skills, and Skill Cov. denotes the fraction of available skills used at least once. Comp. and Raw denote compressed policy decisions and expanded atomic calls, respectively. The reported Avg.~Comp./Raw is the arithmetic mean of the six setting-level ratios, rather than the ratio between the averaged Comp. and Raw counts.

\begin{table*}[ht]
\centering
\scriptsize
\setlength{\tabcolsep}{3.8pt}
\renewcommand{\arraystretch}{1.08}
\caption{
Additional evaluation metrics of CIPO across benchmarks and backbones.
}
\label{tab:additional_metrics}
\resizebox{\textwidth}{!}{
\begin{tabular}{llcccccccc}
\toprule
Benchmark & Backbone
& Prec. & Rec. & NTAcc & Prog.
& Skill Use & Skill Cov.
& Comp. & Raw \\
\midrule
TOOLATHLON & Qwen-7B
& 45.30 & 28.75 & 19.72 & 83.00
& 16.03 & 29.17
& 6.30 & 7.37 \\

TOOLATHLON & Llama-8B
& 59.50 & 40.70 & 23.67 & 95.00
& 29.24 & 31.25
& 8.55 & 11.07 \\

TOUCAN & Qwen-7B
& 47.80 & 49.33 & 47.80 & 78.00
& 9.50 & 27.50
& 7.30 & 8.49 \\

TOUCAN & Llama-8B
& 49.70 & 51.35 & 50.20 & 84.00
& 22.80 & 42.30
& 8.95 & 11.02 \\

TRAJECT-Bench & Qwen-7B
& 53.40 & 48.80 & 52.15 & 78.00
& 9.10 & 19.15
& 4.55 & 5.00 \\

TRAJECT-Bench & Llama-8B
& 60.05 & 52.22 & 46.24 & 96.00
& 2.04 & 21.28
& 12.25 & 14.23 \\
\midrule
Average & --
& 52.63 & 45.19 & 39.96 & 85.67
& 14.79 & 28.44
& 7.98 & 9.53 \\
\bottomrule
\end{tabular}
}
\end{table*}

The additional metrics provide a more detailed view of CIPO's behavior. The precision, recall, and NTAcc columns decompose the Tool F1 results in Table~\ref{tab:main_results}, showing whether gains come from more accurate tool selection, broader tool coverage, or better step-level choices. The skill-related columns show that CIPO uses composite skills across settings without relying on the entire skill library uniformly. Skill Use varies across benchmarks, with lower usage on TRAJECT-Bench, which is consistent with its weaker decision-compression gain. The Comp. and Raw columns further explain the Avg.Comp./Raw value reported in the main table by showing the compressed policy decisions and the corresponding expanded atomic executions.

\paragraph{Evaluation uncertainty. }
Table~\ref{tab:cipo_ci} reports uncertainty estimates for the main CIPO results. For Tool F1 and Comp./Raw, we estimate 95\% confidence intervals by resampling evaluation episodes with replacement and recomputing the corresponding episode-level metric. TSR is judge-based and is independently evaluated five times on the same set of 100 episodes. We report its mean and standard deviation across these five judge runs. The main table reports the mean TSR for readability, while Table 13 additionally shows the variation across repeated evaluations.

\begin{table}[ht]
\centering
\small
\setlength{\tabcolsep}{4pt}
\renewcommand{\arraystretch}{1.08}
\caption{
Uncertainty estimates for CIPO. Tool F1 and Comp./Raw are reported with episode-level bootstrap 95\% confidence interval half-widths. TSR is reported as mean ± standard deviation across five independent judge evaluations of the same 100 evaluation episodes.
}
\label{tab:cipo_ci}
\begin{tabular}{llccc}
\toprule
Benchmark & Backbone & F1 & TSR & Comp./Raw \\
\midrule
TOOLATHLON & Qwen-7B
& \(35.17\pm1.42\) & \(30.78\pm1.65\) & \(0.855\pm0.024\) \\

TOOLATHLON & Llama-8B
& \(48.32\pm1.87\) & \(40.55\pm2.01\) & \(0.772\pm0.028\) \\

TOUCAN & Qwen-7B
& \(48.55\pm1.62\) & \(35.50\pm1.70\) & \(0.860\pm0.024\) \\

TOUCAN & Llama-8B
& \(50.51\pm1.70\) & \(41.32\pm1.85\) & \(0.812\pm0.027\) \\

TRAJECT-Bench & Qwen-7B
& \(50.99\pm1.45\) & \(32.00\pm1.63\) & \(0.910\pm0.020\) \\

TRAJECT-Bench & Llama-8B
& \(55.86\pm1.44\) & \(38.75\pm1.81\) & \(0.861\pm0.018\) \\
\midrule
Average & --
& \(48.23\pm1.15\) & \(36.48\pm1.35\) & \(0.845\pm0.030\) \\
\bottomrule
\end{tabular}
\end{table}

\section{Dataset Statistics}
\label{app:dataset_stats}

We evaluate CIPO on three tool-use benchmarks that cover complementary sources of difficulty. TOOLATHLON contains diverse long-horizon tasks that require agents to maintain coherent tool-use decisions across many dependent steps. TOUCAN contains tool-agentic trajectories built from real-world MCP environments and emphasizes tool selection from a large API space. TRAJECT-Bench focuses on trajectory-aware tool-use evaluation, where the ordering and coherence of tool invocations are important in addition to the final response. These benchmarks allow us to test whether CIPO improves adaptive tool granularity under long trajectories, large tool pools, and trajectory-sensitive execution.

\begin{table}[ht]
\centering
\small
\caption{Statistics of the processed benchmark splits used in our experiments.}
\label{tab:dataset_stats}
\begin{tabular}{lccc}
\toprule
Benchmark & Avg. tool invocations / episode & \# Unique tools & Main difficulty \\
\midrule
TOOLATHLON & \(\sim 13.5\) & 1,250 & Long-horizon dependency \\
TOUCAN & \(\sim 3.2\) & 15,294 & Large-scale tool selection \\
TRAJECT-Bench & \(\sim 6.4\) & 715 & Trajectory coherence \\
\bottomrule
\end{tabular}
\end{table}

TOOLATHLON has the longest trajectories among the three benchmarks, with about 13.5 tool invocations per episode. It is well suited for evaluating whether composite skills can reduce repeated low-level decisions while preserving the dependencies between distant tool invocations. This setting is important for CIPO because long trajectories provide more opportunities for recurring tool-use routines, but also make incorrect granularity choices more costly.

TOUCAN has shorter trajectories on average, but its tool pool is substantially larger. This benchmark tests whether the policy can identify useful tools and skills from a broad action space. For CIPO, TOUCAN is useful for evaluating whether skill construction and counterfactual granularity training remain effective when the main difficulty comes from tool selection rather than trajectory length alone.

TRAJECT-Bench lies between TOOLATHLON and TOUCAN in trajectory length, with about 6.4 tool invocations per episode. Its trajectory-aware evaluation emphasizes whether the generated tool-use path is coherent, ordered, and aligned with the task structure. This makes it useful for testing whether CIPO improves tool-use quality beyond simply increasing skill invocation.

The three benchmarks are complementary. TOOLATHLON stresses long-horizon dependency, TOUCAN stresses large-scale tool selection, and TRAJECT-Bench stresses coherent tool-use trajectories. This combination allows us to evaluate CIPO across different forms of tool-use difficulty and to test whether adaptive tool granularity selection is useful beyond a single benchmark setting.

\section{Task Success Evaluation}
\label{app:llm_judge}

We use an LLM judge to evaluate task completion because tool-use tasks can often be solved by multiple valid tool trajectories. Exact matching against a single reference trajectory may understate success when the agent uses functionally equivalent tools or follows a different valid order. For each benchmark-backbone-method setting, we evaluate 100 episodes and independently judge the resulting trajectories five times using the same judge model and prompt. The main TSR is the arithmetic mean of the five run-level TSR values.

\paragraph{Judge model and inputs.}
The judge model is Qwen3.5-plus. We set temperature to 0.1 and maximum generation length to 512 tokens. Each evaluation trajectory is judged independently in five runs. For each episode, the input contains four fields: the user task, the ground-truth tool chain, the agent's actual tool-call chain, and the agent's final answer. Each tool invocation is formatted with its step index, tool name, arguments, and tool output. To keep the prompt length controlled, each tool output is truncated to 500 characters, the user instruction is truncated to 1000 characters, and the final response is truncated to 500 characters.

\paragraph{Evaluation principle.}
The judge evaluates the agent's tool-use strategy rather than low-level simulator artifacts. In particular, execution errors returned by the simulator, empty outputs, or API-like failures are not treated as agent errors if the selected tool and arguments are reasonable. The reference tool chain is used as one possible solution path, not as the only valid answer. Functionally equivalent tools, minor naming differences, version suffixes, and format differences are considered acceptable when they satisfy the same task requirement. For long tasks with more than five reference tools, partial completion is allowed when the agent covers the main stages of the task.

\paragraph{TSR computation.}
The judge assigns each episode one of three completion labels: \texttt{yes}, \texttt{partial}, or \texttt{no}. The main tables report Task Success Rate (TSR) as the proportion of valid judgments labeled \texttt{yes} or \texttt{partial}. This metric includes partial completion in long-horizon tool-use tasks. We separately record the \texttt{yes}-only strict success rate, which excludes all episodes labeled \texttt{partial} or \texttt{no}. To assess whether the improvement persists under stricter evaluation, we additionally report repeated \texttt{yes}-only judgments and environment-based task verification in Section~\ref{app:strict_execution_success}.

\begin{promptbox}{Judge System Prompt}
You are a professional evaluator of tool-use trajectories. Your task is to judge whether an AI agent's tool-call strategy is reasonable and whether it shows a correct understanding of the user task.

Important evaluation principles:
\begin{enumerate}
    \item Evaluate the tool-use strategy, not simulator artifacts. Tool errors, empty returns, or API failures should not be penalized when the chosen tool type and arguments are reasonable.
    \item The ground-truth tool chain is only one possible solution path. Functionally equivalent tools or alternative valid APIs should be accepted.
    \item Focus on task decomposition. If the agent correctly identifies the main intent and selects tools in the right direction, it should receive credit even when the exact tool names differ.
    \item Match tool names flexibly. Version suffixes, case differences, underscores and hyphens should be treated as equivalent when their functions match.
    \item Be tolerant of small models. The evaluated agents are 1.5B to 8B models, and selecting relevant tool types in complex multi-step tasks should receive credit.
    \item For complex tasks with more than five reference tools, partial coverage of key stages can be judged as partial completion.
    \item Repeated calls should not be penalized by themselves. They may reflect attempts with different parameters.
\end{enumerate}
\end{promptbox}

\paragraph{Aggregation.}
For each setting, we independently evaluate the same 100 episodes in five judge runs. 
For judge run \(r\), we aggregate outputs over all valid episodes and compute
\begin{equation}
\mathrm{TSR}^{(r)}
=
\frac{N_{\mathrm{yes}}^{(r)}+N_{\mathrm{partial}}^{(r)}}
{N_{\mathrm{valid}}^{(r)}}.
\end{equation}
A judge response that cannot be parsed into the required JSON format is marked as an evaluation error and excluded from \(N_{\mathrm{valid}}^{(r)}\) for that run. 
The TSR reported in the main tables is the mean over the five independent judge runs:
\begin{equation}
\mathrm{TSR}
=
\frac{1}{R}\sum_{r=1}^{R}\mathrm{TSR}^{(r)},
\qquad R=5.
\label{eq:tsr_judge}
\end{equation}
We also report the average judge score, strict success rate, partial rate, and fail rate.

\begin{promptbox}{Judge User Prompt Template}
\textbf{User task}

\texttt{\{user\_prompt\}}

\medskip
\textbf{Reference tool chain}

The following ground-truth tool chain is provided as a reference path only, not as the unique correct solution:

\texttt{\{gt\_tools\}}

\medskip
\textbf{Agent tool-call chain}

\texttt{\{tool\_chain\_detail\}}

\medskip
\textbf{Agent final response}

\texttt{\{final\_response\}}

\medskip
Please evaluate the agent's tool-use strategy and output only the following JSON object:

\begin{verbatim}
{
  "task_completed": "yes/partial/no",
  "score": <integer from 0 to 10>,
  "tool_selection_correct": <true/false>,
  "tool_order_reasonable": <true/false>,
  "key_tools_covered": <number of covered key tools>,
  "key_tools_total": <number of key reference steps>,
  "reason": "<brief reason in 1-2 sentences>"
}
\end{verbatim}

Scoring guideline:
\begin{itemize}
    \item 9--10: the tool-use strategy is correct and covers all key steps.
    \item 7--8: the strategy is correct and covers most key steps with minor omissions.
    \item 5--6: the agent understands the task and covers major stages.
    \item 3--4: the agent calls only a few relevant tools and misses core steps.
    \item 1--2: the tool invocations are mostly unrelated to the task.
    \item 0: no effective tool invocation is made.
\end{itemize}

Completion label:
\begin{itemize}
    \item \texttt{yes}: score \(\geq 7\).
    \item \texttt{partial}: score \(\geq 4\) and score \(<7\).
    \item \texttt{no}: score \(<4\).
\end{itemize}
\end{promptbox}

\subsection{Strict and Execution-Based Task Success}
\label{app:strict_execution_success}

We complement the main TSR with two evaluations to examine whether CIPO's improvement persists when partial labels receive no credit and when task completion is verified directly from the execution environment.

\paragraph{Strict judge-based evaluation.}
In addition to the repeated \texttt{Qwen3.5-plus} evaluation used for the main TSR, we re-evaluate the same 100 episodes in each of the six benchmark--backbone settings using \texttt{gpt-4o}. Method identities are hidden from the judge, the temperature is set to 0.1, and each trajectory is independently judged five times. Only the \texttt{yes} label, corresponding to a score of at least 7, counts as strict success. For each judge run, we average strict success rates across the six settings, and report the mean and standard deviation over the five runs.

\paragraph{Execution-based evaluation.}
We additionally apply TOOLATHLON's original task-specific execution evaluators~\citep{toolathlon} to the same 100 TOOLATHLON tasks for Qwen2.5-7B and Llama3.1-8B over three training seeds. These evaluators check whether the resulting environment state satisfies the task requirements, without relying on an LLM judgment of the tool trajectory. Alternative valid tool paths are accepted when they produce the required final state. For each seed, we average the success rates across the two backbones, then report the mean and standard deviation across the three seeds. This execution-based evaluation is restricted to TOOLATHLON; TOUCAN and TRAJECT-Bench are covered by the strict judge-based evaluation above.

\begin{table}[ht]
\centering
\small
\setlength{\tabcolsep}{8pt}
\renewcommand{\arraystretch}{1.12}
\caption{
Strict and execution-based task success (\%).
The \texttt{yes}-only column averages all six benchmark--backbone settings and reports mean $\pm$ standard deviation across five judge runs.
The execution-based column covers TOOLATHLON and reports mean $\pm$ standard deviation across three training seeds after averaging the two backbones.
}
\label{tab:strict_execution_success}
\begin{tabular}{lcc}
\toprule
Method
& \shortstack{\texttt{yes}-only strict success\\All six settings}
& \shortstack{Execution-based success\\TOOLATHLON} \\
\midrule
SFT
& $16.32 \pm 0.42$
& $11.67 \pm 0.58$ \\
Standard GRPO
& $19.63 \pm 0.51$
& $14.83 \pm 0.76$ \\
GRPO + Skills
& $21.40 \pm 0.47$
& $17.67 \pm 0.58$ \\
\textbf{CIPO}
& $\mathbf{27.68} \pm 0.54$
& $\mathbf{25.17} \pm 0.76$ \\
\bottomrule
\end{tabular}
\end{table}

\paragraph{Results.}
Table~\ref{tab:strict_execution_success} presents the \texttt{yes}-only results. CIPO achieves the highest mean in both columns, exceeding GRPO + Skills by 6.28 percentage points in \texttt{yes}-only strict success and 7.50 percentage points in the TOOLATHLON comparison. The strict evaluation shows that CIPO retains an advantage when partial labels are excluded from the success count. The corresponding advantage on TOOLATHLON provides complementary evidence of improved task completion without relying on LLM judgment or partial-credit

\section{Effect of the Skill-mining Budget}
\label{app:bpe_budget_ablation}

We further analyze the effect of the BPE merge budget \(K\), which controls the maximum number of merge rounds during budget-constrained skill mining. This hyperparameter directly affects the size and granularity of the skill library. A smaller budget produces a compact library, but it may miss reusable tool-use routines that appear across successful trajectories. A larger budget can recover more candidate routines, but it may also introduce redundant, overlapping, or task-specific skills. These redundant skills increase the action space and make it harder for the policy to decide which composite action should replace atomic tools.

\begin{figure}[ht]
    \centering
    \includegraphics[width=0.7\linewidth]{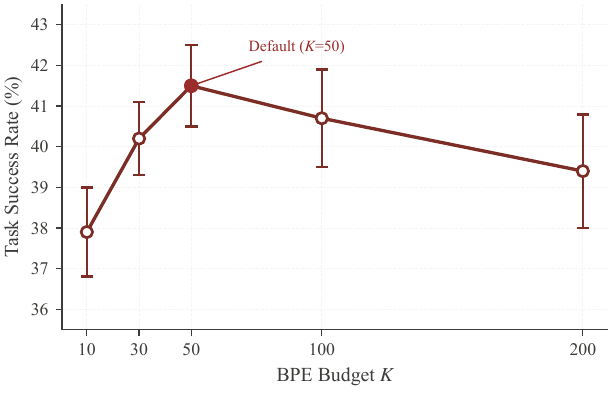}
    \caption{
    Effect of the BPE merge budget \(K\). TSR improves as more reusable routines are mined, but declines when the skill library becomes too large and redundant. The default value \(K=50\) gives the best average TSR.
    }
    \label{fig:bpe_budget_ablation}
\end{figure}

Figure~\ref{fig:bpe_budget_ablation} shows the task success rate under different values of \(K\). TSR increases from \(K=10\) to \(K=50\), indicating that a very small budget under-extracts useful recurring patterns. The gain becomes weaker after \(K=50\), and performance declines when \(K\) is further increased to 100 and 200. This trend suggests that skill mining benefits from a moderate budget: the library should be large enough to cover common routines, but not so large that it creates excessive ambiguity. The default value \(K=50\) achieves the best average TSR, which supports the budgeted design used in the main experiments.

\section{Task-level Decision Compression}
\label{app:decision_compression}

The main table reports Comp./Raw as an aggregate measure of decision compression. To better understand this effect at the task level, we visualize the relation between policy-level decisions and expanded atomic calls for individual evaluation episodes. For each episode, the horizontal axis records the number of decisions made by the policy, where each composite skill counts as one decision. The vertical axis records the number of atomic tool invocations after expanding all executed skills. The diagonal line \(y=x\) indicates no compression: every policy decision corresponds to exactly one atomic operation.

\begin{figure}[ht]
    \centering
    \includegraphics[width=0.5\linewidth]{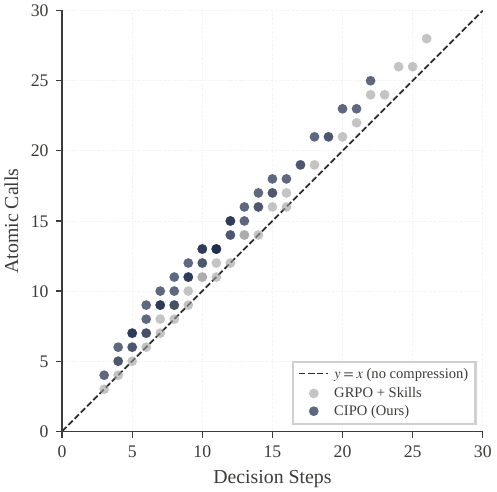}
    \caption{
    Task-level decision compression. Each point represents one evaluation episode. The diagonal line indicates no compression. Compared with GRPO + Skills, CIPO places more episodes above the diagonal, showing that composite skills allow fewer policy-level decisions to execute more atomic operations.
    }
    \label{fig:decision_vs_atomic}
\end{figure}

Figure~\ref{fig:decision_vs_atomic} compares CIPO with GRPO + Skills in the same augmented action space. GRPO + Skills remains close to the diagonal, indicating limited decision compression. In contrast, CIPO places more episodes above the diagonal, where each policy decision corresponds to more than one executed atomic operation. This means that a smaller number of policy decisions can trigger a larger number of executed atomic operations through composite skills.

This task-level pattern complements the Comp./Raw results in Table~\ref{tab:main_results}. The average ratio shows that CIPO reduces policy-level decisions overall, while the scatter plot shows that this reduction is not driven by a few isolated cases. Instead, many episodes exhibit moderate compression, and some longer episodes show larger gaps between decision steps and atomic calls. These results support the claim that CIPO improves efficiency by learning when composite skills can safely replace repeated atomic tool choices.

\section{SFT Initializations and GRPO Baselines}
\label{app:grpo_baselines}

\paragraph{Separate SFT initializations.}
We use separate SFT initializations for atomic-only and skill-enabled policies. Standard GRPO starts from an SFT model trained with atomic tools only. GRPO + Skills starts from a separately trained SFT model with access to both atomic tools and composite skills. CIPO also follows the skill-enabled SFT pipeline described in Section~3.4. The SFT baseline reported in the ablation tables and skill-usage analyses denotes the skill-enabled model.

\paragraph{Action spaces and policy optimization.}
Standard GRPO uses the atomic action space $\mathcal{A}_{\mathrm{atom}}$ during both reinforcement learning and evaluation. GRPO + Skills uses the augmented action space $\mathcal{A} = \mathcal{A}_{\mathrm{atom}} \cup \mathcal{A}_{\mathrm{skill}}$ and applies the standard GRPO objective without counterfactual branching or paired reward adjustments. CIPO uses the same augmented action space but constructs matched counterfactual branches and adds a bounded supplementary reward based on paired outcome differences.

\paragraph{Interpretation of comparisons.}
Standard GRPO provides an atomic-only baseline, whereas GRPO + Skills provides the skill-enabled baseline for assessing CIPO beyond access to executable skills. The comparison between Standard GRPO and GRPO + Skills includes both the action-space extension and the corresponding SFT preparation. In the ablation tables, $\Delta$F1 and $\Delta$TSR are measured relative to the reported skill-enabled SFT baseline. For Standard GRPO, these differences compare the final policies and do not measure improvement over its own SFT initialization. Analyses of skill invocation and granularity selection use GRPO + Skills as the GRPO baseline.

\paragraph{Decision compression.}
For Standard GRPO, each tool decision corresponds to one atomic call. Comp./Raw therefore equals 1 for trajectories with at least one tool call, and the corresponding decision reduction is 0\%. Skill-enabled methods can invoke composite skills, whose internal atomic calls are expanded when computing Raw.

\section{Baseline Implementation Details}
\label{app:baseline_details}

\begin{table}[ht]
\centering
\small
\setlength{\tabcolsep}{4.8pt}
\renewcommand{\arraystretch}{1.08}
\caption{
Summary of baseline adaptation. ``Prompt'' means that the method injects memory, workflow, or skill descriptions into the prompt. ``Action'' means that the method exposes composite tools or skills as callable actions.
}
\label{tab:baseline_adaptation}
\resizebox{\linewidth}{!}{
\begin{tabular}{lccc}
\toprule
Method group & Built from training trajectories & Used at inference as & Comp./Raw reported \\
\midrule
General tool-use agents & no & atomic tool policy & no \\
Memory and workflow reuse & yes & prompt guidance & no \\
Tool composition & yes & composite action or compiled interface & yes \\
Skill construction & yes & prompt skill or composite action & yes \\
CIPO & yes & executable composite skills + trained policy & yes \\
\bottomrule
\end{tabular}
}
\end{table}

This appendix describes how we adapt baselines to the same tool-use environment. All methods use the same train/test split, backbone models, atomic tool definitions, simulator, maximum turn budget, decoding setting, and output parser. Methods that require memory, workflows, or skills build their auxiliary resources only from training trajectories. Test episodes are never used for memory construction, workflow induction, skill mining, or hyperparameter selection.

\paragraph{Unified evaluation interface.}
All baselines are evaluated as multi-step tool-use agents. At each step, the model receives the task, previous tool calls, tool observations, and available tool specifications, then outputs the next tool call in the same structured JSON format. The generated tool call is executed by the same simulator used by CIPO. For methods without executable composite actions, Comp./Raw equals 1 for trajectories with at least one tool call. Dashes in the main comparison table indicate that this value is not reported.

\paragraph{General tool-use baselines.}
ReAct is implemented as an atomic-tool agent that interleaves reasoning, tool calls, and final answers. ToolLLM is adapted as a tool-retrieval and atomic invocation baseline over the same candidate tool schema. Both methods operate over atomic tools only and do not construct memory, workflows, or composite skills.

\paragraph{ExpeL.}
We implement an ExpeL-style dual-memory baseline. Episodic memory stores training trajectories with task descriptions, tool sequences, arguments, observations, and success labels, mainly indexing successful trajectories for retrieval. Insight memory stores natural-language tool-use rules extracted by comparing successful and failed trajectories within the same task group. At evaluation time, retrieved episodic examples and ranked insight rules are inserted into the system prompt. ExpeL therefore provides both concrete examples and abstract guidance, but it does not change the action space.

\paragraph{MemP.}
MemP is implemented as a procedural-memory baseline. Each training episode is distilled into a natural-language workflow with its task query, task name, and tool sequence. Both successful and failed samples are used; failed samples are converted through a corrective prompt into more appropriate workflows. At evaluation time, the current task retrieves the top \(k=3\) workflows by query similarity, and these workflows are injected into the prompt together with a reduced candidate tool set. MemP guides atomic tool selection through retrieved procedures.

\paragraph{Mem0.}
Mem0 is adapted as a consolidated long-term procedural memory. Each training episode is summarized into a short reusable workflow and an associated tool sequence. When a new memory is produced, similar existing memories are retrieved and an LLM decides whether to add, update, or skip the new item. At evaluation time, the system retrieves relevant procedure memories once at the beginning of each episode and keeps them fixed throughout the interaction. The retrieved memories guide tool selection but remain prompt-level knowledge.

\paragraph{AWM and AutoManual.}
For AWM, successful training episodes are grouped by task type and summarized into workflow blocks containing a workflow name, applicability condition, and ordered step templates. AutoManual similarly constructs instruction-style manuals that describe stable tool-use procedures and parameter hints. During evaluation, relevant workflows or manuals are retrieved and inserted into the prompt. Both methods reuse procedural knowledge as guidance rather than as executable actions.

\paragraph{Tool composition baselines.}
AWO is adapted by identifying repeated workflow traces from successful training trajectories and registering them as meta-tools when their internal tool sequence and interface can be determined. LLMCompiler is adapted as a planning and function-calling baseline that groups independent calls into execution stages and executes dependent stages sequentially in our simulator. LLM-Tool Compiler is adapted by exposing fused interfaces for similar or frequently co-occurring tools. These baselines introduce higher-level interfaces or compiled execution structures, but they do not train atomic-versus-skill choices with matched counterfactual branches.

\paragraph{Skill construction baselines.}
SkillRL is adapted as a skill-bank baseline that distills training trajectories into reusable skill descriptions and, when possible, exposes concrete tool subsequences as callable composite actions. SKILL0 is adapted as a skill-internalization baseline, where skill descriptions are provided during training and gradually reduced so that the policy internalizes their use. Trace2Skill extracts trajectory-local lessons from training traces and consolidates them into a skill directory; retrieved skills are used as prompt guidance, and executable subsequences are exposed as composite actions when their schema can be inferred. These methods reuse skills, but they do not provide CIPO's first-eligible branch comparison or paired reward difference for granularity selection.

\paragraph{Internal GRPO baselines.}
SFT denotes the skill-enabled supervised baseline. Standard GRPO starts from a separate atomic-only SFT model and uses only atomic tools during reinforcement learning and evaluation. GRPO + Skills starts from the skill-enabled SFT model and uses the augmented action space without counterfactual branching. Appendix~\ref{app:grpo_baselines} details these configurations. 
The w/o Budgeted Mining variant replaces constrained mining with a less controlled skill library, while keeping the rest of the pipeline unchanged. CIPO-random uses the same group size \(G=4\) as CIPO-first, but samples the branch point from all eligible granularity decisions instead of using the first eligible decision.

\paragraph{Fairness considerations.}
All baselines are compared under the same model backbones, evaluation tasks, tool simulator, decoding budget, and metric definitions. Memory and workflow methods may use training trajectories, but only as prompt-level guidance. Composite-action methods may register additional callable actions, but their internal executions are expanded when computing Raw. This setup ensures that differences in performance reflect the method-specific use of memory, workflow reuse, tool composition, skill construction, or granularity training rather than differences in the evaluation environment.

\section{Broader Impact}
\label{app:broader_impact}

CIPO studies how LLM agents choose between atomic tools and composite skills during long-horizon tool use. The main positive impact is improved efficiency and reliability for tool-use agents. By compressing recurring tool-use routines into executable skills, CIPO can reduce repeated low-level decisions and help agents complete complex workflows with fewer policy-level actions. This can make tool-use systems more practical in settings such as research assistance, software workflows, document processing, and API-based data analysis.

The same capability also requires care. More effective tool-use agents may execute longer or more complex action sequences, and errors in tool selection can propagate through downstream tool calls. Composite skills may also hide intermediate steps if they are implemented as opaque pipelines. CIPO mitigates this issue by making skills executable, trace-aware, and interruptible: each internal step can be recorded, inspected, and interrupted when execution fails. The counterfactual reward is also bounded, which reduces the risk that the policy over-optimizes for skill invocation at the cost of task completion.

Our experiments are conducted in benchmarked tool environments rather than deployed high-stakes settings. We do not evaluate or recommend direct deployment in domains where erroneous tool execution could cause material harm, such as medical, legal, financial, security-critical, or safety-critical decision making. In such settings, CIPO-style agents should be combined with human approval, sandboxed execution, tool permission control, logging, and domain-specific safety checks. Future work can further study online monitoring, dynamic tool updates, and safeguards for preventing unsafe or unintended composite skill execution.

\appendix
\end{document}

%% file: macro.tex
\newcommand{\x}{\mathbf{x}}

\definecolor{blanchedalmond}{rgb}{1.0, 0.92, 0.8}
\definecolor{carmine}{rgb}{0.59, 0.0, 0.09}
\definecolor{lightblue}{rgb}{0.22,0.45,0.70}%

\renewcommand{\mathbf}{\boldsymbol}

\makeatletter
\def\Ddots{\mathinner{\mkern1mu\raise\p@
\vbox{\kern7\p@\hbox{.}}\mkern2mu
\raise4\p@\hbox{.}\mkern2mu\raise7\p@\hbox{.}\mkern1mu}}
\makeatother

\definecolor{amaranth}{rgb}{0.9, 0.17, 0.31}
\definecolor{antiquebrass}{rgb}{0.8, 0.58, 0.46}
\definecolor{antiquefuchsia}{rgb}{0.57, 0.36, 0.51}
\definecolor{chromeyellow}{rgb}{0.31, 0.47, 0.26}